%% file: main.tex
\documentclass[sigconf]{acmart}
\AtBeginDocument{%
  \providecommand\BibTeX{{%
    \normalfont B\kern-0.5em{\scshape i\kern-0.25em b}\kern-0.8em\TeX}}}

\usepackage{algorithm}
\usepackage{algorithmic}
\input{math_command.tex}

\usepackage{multirow}
\usepackage{latexsym}
\usepackage{url}
\usepackage{amsmath,nccmath}
\usepackage{xcolor}
\usepackage{graphicx}
\usepackage{caption}
\usepackage{enumitem}
\usepackage{array}

\usepackage{subfigure}
\newcolumntype{P}[1]{>{\centering\arraybackslash}p{#1}}

\newcommand{\method}{SDIL}

\copyrightyear{2023}
\acmYear{2023}
\setcopyright{acmlicensed}\acmConference[KDD '23]{Proceedings of the 29th ACM SIGKDD Conference on Knowledge Discovery and Data Mining}{August 6--10, 2023}{Long Beach, CA, USA}
\acmBooktitle{Proceedings of the 29th ACM SIGKDD Conference on Knowledge Discovery and Data Mining (KDD '23), August 6--10, 2023, Long Beach, CA, USA}
\acmPrice{15.00}
\acmDOI{10.1145/3580305.3599506}
\acmISBN{979-8-4007-0103-0/23/08}

\begin{document}

\title{Skill Disentanglement for Imitation Learning from Suboptimal Demonstrations}

% \renewcommand{\shortauthors}{Trovato and Tobin, et al.}

%\title[Skill Disentanglement for Imitating Suboptimal Demonstrations]{Skill Disentanglement for Imitation Learning from Suboptimal Demonstrations}

\author{Tianxiang Zhao}
\affiliation{%
  \institution{the Pennsylvania State University}
  \state{Pennsylvania}
  \country{USA}
}
\email{tkz5084@psu.edu}

\author{Wenchao Yu}
\affiliation{%
  \institution{NEC Laboratories America}
  \state{New Jersey}
  \country{USA}
}
\email{wyu@nec-labs.com}

\author{Suhang Wang}
\affiliation{%
  \institution{the Pennsylvania State University}
  \state{Pennsylvania}
  \country{USA}
}
\email{szw494@psu.edu}

\author{Lu Wang}
\affiliation{%
  \institution{East China Normal University}
  \state{Shanghai}
  \country{China}
}
\email{luwang@stu.ecnu.edu.cn}

\author{Xiang Zhang}
\affiliation{%
  \institution{the Pennsylvania State University}
  \state{Pennsylvania}
  \country{USA}
}
\email{xzz89@psu.edu}

\author{Yuncong Chen}
\affiliation{%
  \institution{NEC Laboratories America}
  \state{New Jersey}
  \country{USA}
}
\email{yuncong@nec-labs.com}

\author{Yanchi Liu}
\affiliation{%
  \institution{NEC Laboratories America}
  \state{New Jersey}
  \country{USA}
}
\email{yanchi@nec-labs.com}

\author{Wei Cheng}
\affiliation{%
  \institution{NEC Laboratories America}
  \state{New Jersey}
  \country{USA}
}
\email{weicheng@nec-labs.com}

\author{Haifeng Chen}
\affiliation{%
  \institution{NEC Laboratories America}
  \state{New Jersey}
  \country{USA}
}
\email{haifeng@nec-labs.com}

\renewcommand{\shortauthors}{Tianxiang Zhao et al.}
%% No italics and no comma
%% If needed use a foot or author note to identify equal contribution

\begin{abstract}
Imitation learning has achieved great success in many sequential decision-making tasks, in which a neural agent is learned by imitating collected human demonstrations. However, existing algorithms typically require a large number of high-quality demonstrations that are difficult and expensive to collect. Usually, a trade-off needs to be made between demonstration quality and quantity in practice. Targeting this problem, in this work we consider the imitation of sub-optimal demonstrations, with both a small clean demonstration set and a large noisy set. Some pioneering works have been proposed, but they suffer from many limitations, e.g., assuming a demonstration to be of the same optimality throughout time steps and failing to provide any interpretation w.r.t knowledge learned from the noisy set. Addressing these problems, we propose {\method} by evaluating and imitating at the sub-demonstration level, encoding action primitives of varying quality into different skills. Concretely, {\method} consists of a high-level controller to discover skills and a skill-conditioned module to capture action-taking policies, and is trained following a two-phase pipeline by first discovering skills with all demonstrations and then adapting the controller to only the clean set. A mutual-information-based regularization and a dynamic sub-demonstration optimality estimator are designed to promote disentanglement in the skill space.  Extensive experiments are conducted over two gym environments and a real-world healthcare dataset to demonstrate the superiority of {\method} in learning from sub-optimal demonstrations and its improved interpretability by examining learned skills. 

%learning a hierarchical framework, with a high-level controller module to encode skills and a skill-conditioned action module to capture action primitives. The latent generation of the clean and noisy demonstrations would be modeled with overlapped skill selection in the disentangled skill space. This design enables efficient utilization of the noisy demonstration set while preventing its negative influence as those low-quality segments may select different skills. Concretely, we design a two-phase framework by first discovering the skills with both demonstration sets and then adapting them to only the clean set. A mutual-information-based regularization is designed to promote the disentanglement of skills. Furthermore, a dynamic algorithm is proposed to incorporate latent similarities and estimated optimality of segments into the skill discovery phase, which will be updated along with training. We conduct extensive experiments over two gym environments and a real-world healthcare dataset to demonstrate the superiority of {\method} in learning from sub-optimal demonstrations and its improved interpretability by examining learned skills.

\end{abstract}



%%
%% Keywords. The author(s) should pick words that accurately describe
%% the work being presented. Separate the keywords with commas.

\maketitle

\input{secs/introduction}

\input{secs/related_work}

\input{secs/method}

\input{secs/experiment}

\input{secs/conclusion}

\section{Acknowledgments}
This material is based upon work partially supported by National Science Foundation (NSF) under grant number IIS-1909702 and the Army Research Office (ARO) under grant number W911NF21-1-0198 to Suhang Wang.

\newpage
\bibliographystyle{IEEEtran}
\balance
\bibliography{file1}

\newpage
\appendix
\input{secs/appendix.tex}

\end{document}

%% file: math_command.tex
\usepackage{amsmath,amsfonts,bm,nccmath}

\def\eqref#1{equation~\ref{#1}}
\def\1{\bm{1}}

\def\va{{\bm{a}}}

\def\vs{{\bm{s}}}

\def\vz{{\bm{z}}}

\DeclareMathAlphabet{\mathsfit}{\encodingdefault}{\sfdefault}{m}{sl}
\SetMathAlphabet{\mathsfit}{bold}{\encodingdefault}{\sfdefault}{bx}{n}

\newcommand{\E}{\mathbb{E}}

\newcommand{\R}{\mathbb{R}}

\DeclareMathOperator*{\argmax}{arg\,max}

%% file: secs/introduction.tex
\section{Introduction}
Imitation learning (IL), which aims to imitate human demonstrations, has achieved great success in many sequential decision-making tasks, such as assistive robots~\cite{kiran2021deep,hemminghaus2017towards} and human-computer interactions~\cite{zhao2020balancing,huang2021transmart}.  
Generally, IL assumes access to a collected human demonstration set and learns the action policy by mimicking the latent generation process of those demonstrations~\cite{pomerleau1998autonomous,fu2018learning,ho2016generative}. Each demonstration is a sequence of transitions (state-action pairs), generated by experts in the target task. However, the success of most existing IL algorithms crucially depends on the availability of a large and high-quality demonstration set~\cite{pinto2016supersizing}; while in many real-world cases such as autonomous driving and healthcare treatments, it is challenging and expensive to collect them. Furthermore, humans often make mistakes due to various reasons such as difficulty of the task and partial observability~\cite{xu2022discriminator}. The widely-adopted crowd-sourced data collection pipeline would also inevitably lead to demonstrations of varying levels of expertise~\cite{mandlekar2019scaling}.  %a trade-off often needs to be made on the optimality and quantity of collected demonstrations. 
The difficulty of obtaining large-scale high-quality demonstrations challenges many existing IL methods.

Therefore, in this work, we consider a more realistic setting, i.e., we have access to a small clean demonstration set in the accompany of a large noisy demonstration set. In the clean set, demonstrations can be trusted to follow an optimal action-taking policy. On the contrary, the quality of a noisy demonstration can not be guaranteed and it may contain segments (consecutive transitions) of sub-optimal or even poor action selections. For example, in the healthcare domain (Fig.~\ref{fig:motivate}(a)), each demonstration contains the sequential medical treatment history of a particular patient with physiological features as states and medical treatments as actions. The clean set contains carefully examined records that are accurate and effective; while the noisy set is collected in the wild without expert scrutiny. Some noisy demonstrations could contain misdiagnosing, inappropriate use of medications, or dose miscalculations at certain stages. Directly incorporating the noisy set in learning may pollute the dataset and mislead the neural agent, resulting in a negative effect. However, optimal demonstrations are limited in quantity and could be insufficient for successful imitation learning. Hence, it is important to extract useful information from noisy demonstration to help train better IL models.

Extracting useful information from noisy demonstrations is non-trivial, as the quality of actions taken at each time step is not provided. There are some initial efforts on imitating sub-optimal demonstrations~\cite{zhang2021confidence,xu2022discriminator,beliaev2022imitation}. For example, CAIL~\cite{zhang2021confidence} uses meta-learning to estimate the confidence over each noisy demonstration, DWBC~\cite{xu2022discriminator} trains an additional discriminator to distinguish expert and non-expert demonstrations. However, all of these methods are coarse-grained, i.e., them conduct a trajectory-level evaluation and assign the same optimality score to actions across all time steps of the same demonstration~\cite{zhang2021confidence,xu2022discriminator}. This is a strong assumption and limits their effectiveness in real-world applications. For instance, a maze-solving demonstration could contain both proper/efficient and redundant/inefficient segments simultaneously. Furthermore, previous methods lack interpretability. It is difficult to understand what they learn from clean and noisy demonstrations respectively, making the utilization of noisy demonstrations difficult to trust. 

Addressing these challenges, we propose to learn a bi-level skill hierarchy from demonstrations to capture the variations of action policies. Each skill encodes a specific action-taking strategy in a part of the state-space~\cite{sharma2019dynamics}, and each state-action pair of collected demonstrations can be deemed to be generated following a particular skill, as the example in Fig.~\ref{fig:motivate}(b). In modeling the distribution of demonstrations, high-quality segments of noisy demonstrations would have overlapped skill selections with clean demonstrations and can help improve the learning of those selected skills in turn. %\suhang{it would be better if you can use an example to introduce what is hierarchy and how it can help extract useful segments from noisy demonstrations}
 With this design, a set of skills of varying optimality can be extracted from modeling the distribution of demonstrations, and their optimality can be evaluated by analyzing skill-selection behaviors. Furthermore, each skill can be analyzed based on agent behaviors conditioned on it. By examining skill-selection distributions and comparing action-taking policies across skills, stronger interpretability w.r.t learned action primitives will be offered.

\begin{figure}[t]
    \centering
    \subfigure[]{\includegraphics[width=0.46\linewidth]{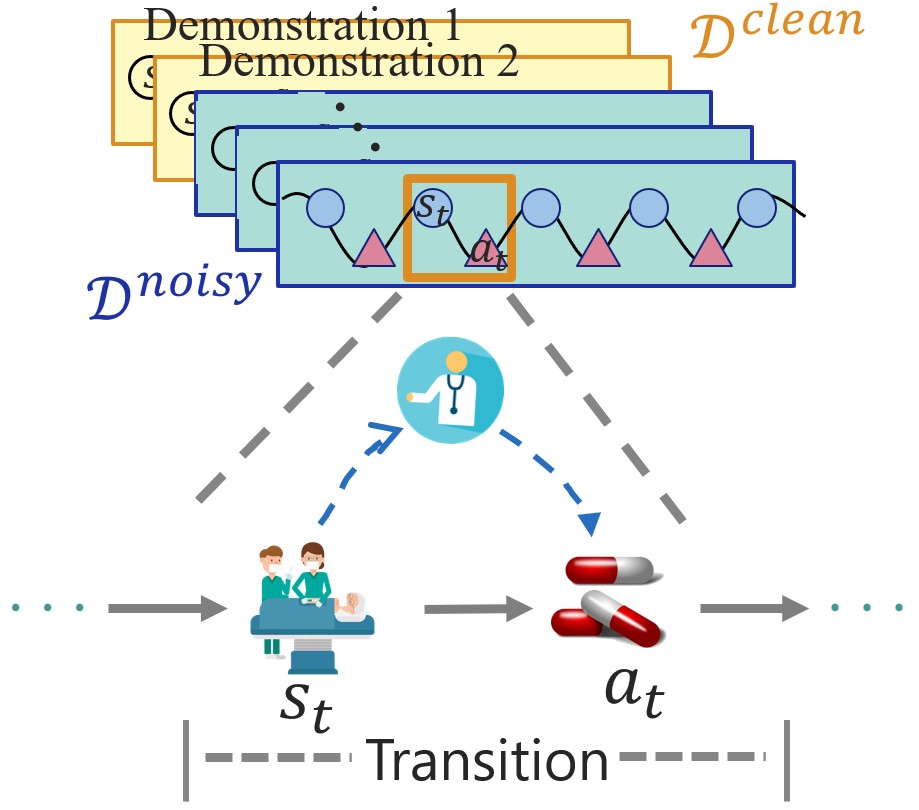}}
    \subfigure[]{\includegraphics[width=0.45\linewidth]{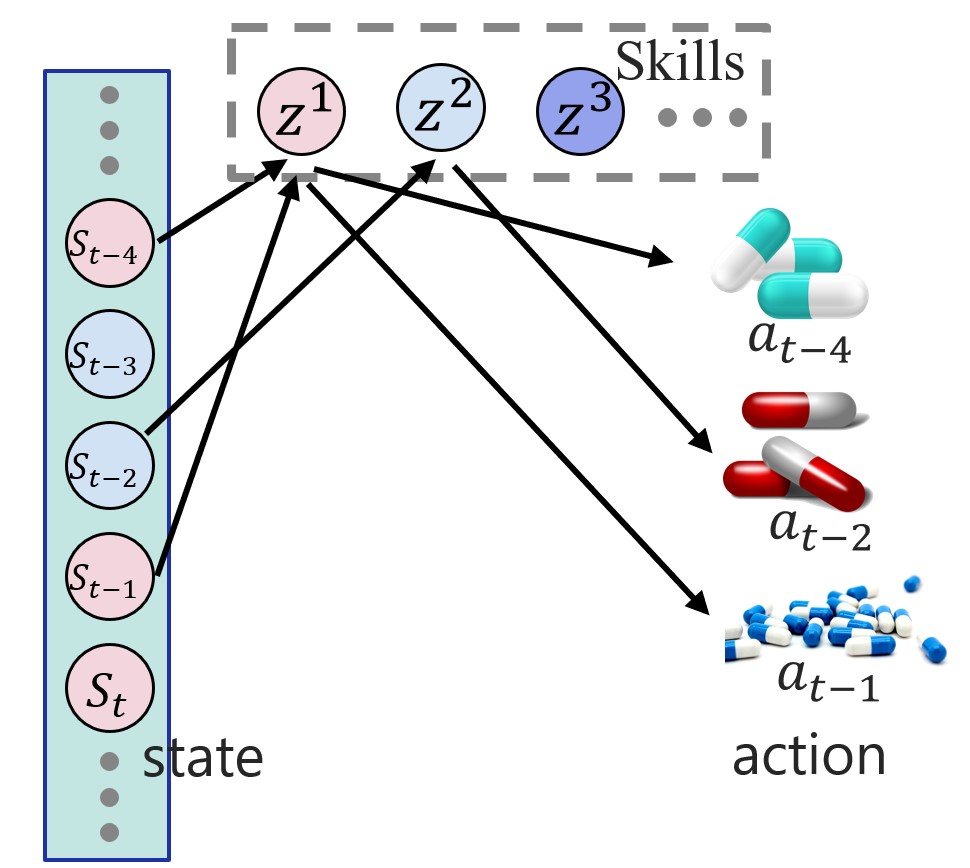}}
    \vskip -1.5em
    \caption{(a) An example in the healthcare domain. We have a small number of clean demonstrations and a large number of noisy demonstrations, each demonstration is composed of a sequence of transitions (state-action pairs). (b) High-level visualization of the skill hierarchy. Each skill encodes a mapping from states to actions. We can identify consecutive transitions modeled by the same skill as a segment with similar level of optimality. }
    \vskip -1em
    \label{fig:motivate}
\end{figure}

\iffalse
\begin{figure}[t]
    \centering
    \includegraphics[width=0.96\linewidth]{figs/example_skill.jpg}
    \vskip -1.5em
    \caption{(a) An example in the healthcare domain. We have a small number of clean demonstrations and a large number of noisy demonstrations, each demonstration can be split into many segments. (b) High-level visualization of the skill hierarchy. Each skill encodes a mapping from states to actions and each segment (shown in color) can be modeled by a particular skill. }
    \vskip -1em
    \label{fig:motivate}
\end{figure}
\fi

Concretely, we design a framework {\method} that comprises a high-level controller to encode observed states and make skill selections, and a low-level action predictor conditioned on the selected skill. {\method} is optimized in a two-phase pipeline. In the first phase, we discover latent skills from both clean and noisy demonstrations, with particularly designed regularization to promote skill disentanglement. With skills learned, in the second phase, we adapt this hierarchical framework by imitating only the clean demonstrations in order to compose an expert-level neural agent. Our codes are available on Github~\footnote{https://github.com/TianxiangZhao/ImitationNoisyDemon}. Our main contributions are:
\begin{itemize}[leftmargin=0.2in]
    \item We study a novel problem of imitation learning with sub-optimal demonstrations by extracting useful information from high optimality segments of noisy demonstrations.%This work considers imitation learning with sub-optimal demonstrations. Especially, we argue that different segments of the same demonstration may be of varying optimality.
    %by discovering latent skills behind its generation. Informative segments of the noisy demonstrations would follow some expert skills, and be utilized to improve the imitation learning process.
    \item We propose a novel framework {\method} which can extract a set of skills from sub-optimal demonstrations and utilize them to compose an expert-level neural agent.%\suhang{which can extract xxx and utlize xxx to xxx???}%a high-level skill controller to encode the skills and a low-level action module to conduct skill-conditioned action prediction.
    %\item To guide the extraction of skills, we design a mutual-information-based regularization, deep clustering, and optimality estimation with Positive-Unlabeled (PU) learning to encourage the discovery of a disentangled skill space.
    \item Experiments are conducted on both synthetic and real-world datasets. Besides quantitative comparisons with existing algorithms, we also provide a set of case studies to analyze the behavior of {\method} and learned skills.
\end{itemize}

%% file: secs/related_work.tex
\section{Related Work}

\subsection{Imitation Learning}
Imitation learning is a special case of reinforcement learning,  which relies on expert trajectories composed of state-action pairs without availability of the interactive environment. Existing IL approaches can be folded into three paradigms, behavior cloning (BC), inverse reinforcement learning (IRL), and generative adversarial imitation learning (GAIL). BC~\cite{pomerleau1998autonomous}  tackles this problem in a supervised manner, learning the action policy as a conditional distribution $\pi(\cdot \mid \vs)$ over actions. Works have observed that BC has no inferior performance compared to other popular IL algorithms such as GAIL~\cite{ho2016generative} with expert demonstrations available~\cite{ma2020adversarial,wang2020adversarial}. IRL~\cite{ziebart2008maximum} dedicates to recover the reward function which would be maximized by expert demonstrations, and learn the agent through exploration over the learned reward. Adversarial  GAIL~\cite{ho2016generative} takes an adversarial approach, train the policy model to generate trajectories that are indistinguishable from expert demonstrations. Adversarial inverse reinforcement learning (AIRL)~\cite{fu2018learning} extends GAIL to simultaneously learn the reward function and the optimal policy, aiming to find the saddle point of a min-max optimization problem. However, these approaches take all demonstrations as equal and assume them to be optimal. They cannot learn well in the existence of sub-optimal demonstrations. 

Some existing works extend traditional IL algorithms to cope with imperfect demonstrations~\cite{yang2021trail,sasaki2020behavioral}. However, most of these methods require prior domain knowledge, like rankings of demonstrations~\cite{sugiyama2012preference,wirth2016model} and demonstration confidence~\cite{wu2019imitation}. CAIL~\cite{zhang2021confidence} utilizes a meta-learning framework to learn from sub-optimal demonstrations with a confidence estimator. In the inner-update step, it conducts a weighted imitation learning based on estimated confidence, and the estimator would be updated in the outer loop. DWBC~\cite{xu2022discriminator} designs a discriminator to evaluate demonstration quality, and ILEED~\cite{beliaev2022imitation} proposes to estimate expertise of demonstrators. However, all these methods conduct a coarse-grained estimation, assuming all time steps of the same demonstration share the same optimality. Different from them, we propose to break each demonstration down and utilize those noisy demonstrations in a fine-grained manner.

\subsection{Hierarchical Reinforcement Learning}
Hierarchical reinforcement learning (HRL) decomposes the full control policy into multiple macro-operators or abstractions, each encoding a short-term decision process~\cite{dayan1992feudal,sutton1999between,vezhnevets2017feudal,bacon2017option}. This hierarchical structure provides intuitive benefits for easier learning and long-term decision-making, as the policy is organized along the hierarchy of multiple levels of abstractions. Typically, the higher-level policy provides conditioning variables~\cite{eysenbach2018diversity,campos2020explore,sharma2019dynamics} or selected sub-goals (options)~\cite{klissarov2021flexible,veeriah2021discovery} to control the behavior of lower-level policy models. Another branch of HRL methods design a hierarchy over the actions to reduce the search space with prior domain knowledge~\cite{sutton1999between,parr1997reinforcement}.

A variety of discussions have been made over the trade-off between generality (wide applicability) and specificity (benefits for specific tasks) of low-level policies~\cite{veeriah2021discovery,klissarov2021flexible,ren2022semi}. Recently, a particular advantage of introducing HRL has also been identified for improved exploration capabilities~\cite{gehring2021hierarchical,jong2008utility}. In this work, we focus primarily on modeling the latent decision policies of sub-optimal demonstrations with a HRL-based framework, using a diverse skill set to model action primitives of varying optimality behind collected demonstrations. As far as we know, we are the first to consider hierarchical skill discovery for automatically learning from noisy demonstrations.

%% file: secs/method.tex
\section{Problem Setup}

In imitation learning, we aim to learn a policy $\pi_\theta$ from the collected demonstration set. Each demonstration $\tau$ is a trajectory, a sequence of transitions (state-action pairs): $\tau = (\vs_0, \va_0, \vs_1, \va_1, ... )$, with $\vs_t \in \mathcal{S}$ and $\va_t \in \mathcal{A}$ being the state and action at time step $t$. $\mathcal{S}$ is the state space and $\mathcal{A}$ is the set of actions.  A policy $\pi: \mathcal{S} \times \mathcal{A} \rightarrow [0,1]$ maps the observed state to a probability distribution over actions. Most existing imitation learning works~\cite{xu2022discriminator,zhang2021confidence} assume that these demonstrations are all optimal, collected from domain experts. However, it is usually challenging to collect a sufficiently large expert demonstration set. Hence, we often need to learn from noisy demonstrations, which challenge existing works.

Specifically, we have a small clean demonstration set $\mathcal{D}^{clean} = \{\tau_i\}_{i=1}^{n_{e}}$ which is drawn from the expert optimal policy $\pi_e$, and a noisy demonstration set $\mathcal{D}^{noisy} = \{\tau_i\}_{i=1}^{n_{o}}$ generated from some other behavioral policies. Note that the qualities of policies used by $\mathcal{D}^{noisy}$ are not evaluated, they could be of similar or much worse than the expert policy $\pi_e$. In this task, we want to learn a policy agent $\pi_{\theta}$ by extracting useful information from both optimal demonstrations $\mathcal{D}^{clean}$ and noisy demonstrations $\mathcal{D}^{noisy}$. 

We assume that these demonstrations are generated from semantically meaningful skills, with each skill encoding a specific action primitive, a sub-policy. For example in the healthcare domain, each skill could represent a strategy of adopting treatment plans in the observance of several particular symptoms. Demonstrations in $\mathcal{D}^{noisy}$ can be split into multiple segments, and useful information can be extracted from segments that are generated from high-quality skills.  Concretely, the task can be formalized as:

\vspace{0.5em}
\noindent{}\textit{Given a small clean demonstration set $\mathcal{D}^{clean}$ and a large noisy demonstration $\mathcal{D}^{noisy}$, we aim to learn a policy agent $\pi_{\theta}$ for action prediction based on observed states: $\pi_{\theta}: \mathcal{S} \times \mathcal{A} \rightarrow [0,1]$ 
}

\section{Skill-based IL from Imperfect Demonstrations}
In this work, we propose to learn from sub-optimal demonstrations with a hierarchical framework, containing both a high-level policy for selecting skills and a low-level policy for action prediction. The high-level policy will maintain a skill set and select the to-be-used skills from observed states, and the low-level policy will decide on actions conditioned on the skill. This framework enables the automatic discovery of skills for utilizing sub-optimal demonstrations. To encourage the disentanglement of skills and obtain a well-performing agent $\pi_{\theta}$, a two-phase framework is proposed: (1) the skill discovery phase utilizing $\mathcal{D}^{clean} \cup \mathcal{D}^{noisy}$ to extract and refine a skill set of varying optimality; (2) the skill reusing phase that adapts learned skills to imitate $\mathcal{D}^{clean}$, transferring the knowledge to learn expert policy $\pi_{\theta}$. In this section, we will first present the hierarchical model architecture, and then introduce the learning of each phase one by one. 

%There are many domains or practical scenarios which the present invention is applicable to. Healthcare domain is one of the examples. In general, in healthcare domain, the sequential medical treatments history of a patient is one expert demonstration. State variables include health records and symptoms, and actions are the usage of treatments. Those demonstrations that patients are fully recovered can be used as expert demonstrations and others can be taken as noisy demonstrations. 

\subsection{Hierarchical Policy for Skill Discovery}

\begin{figure}
    \centering
    \includegraphics[width=\linewidth]{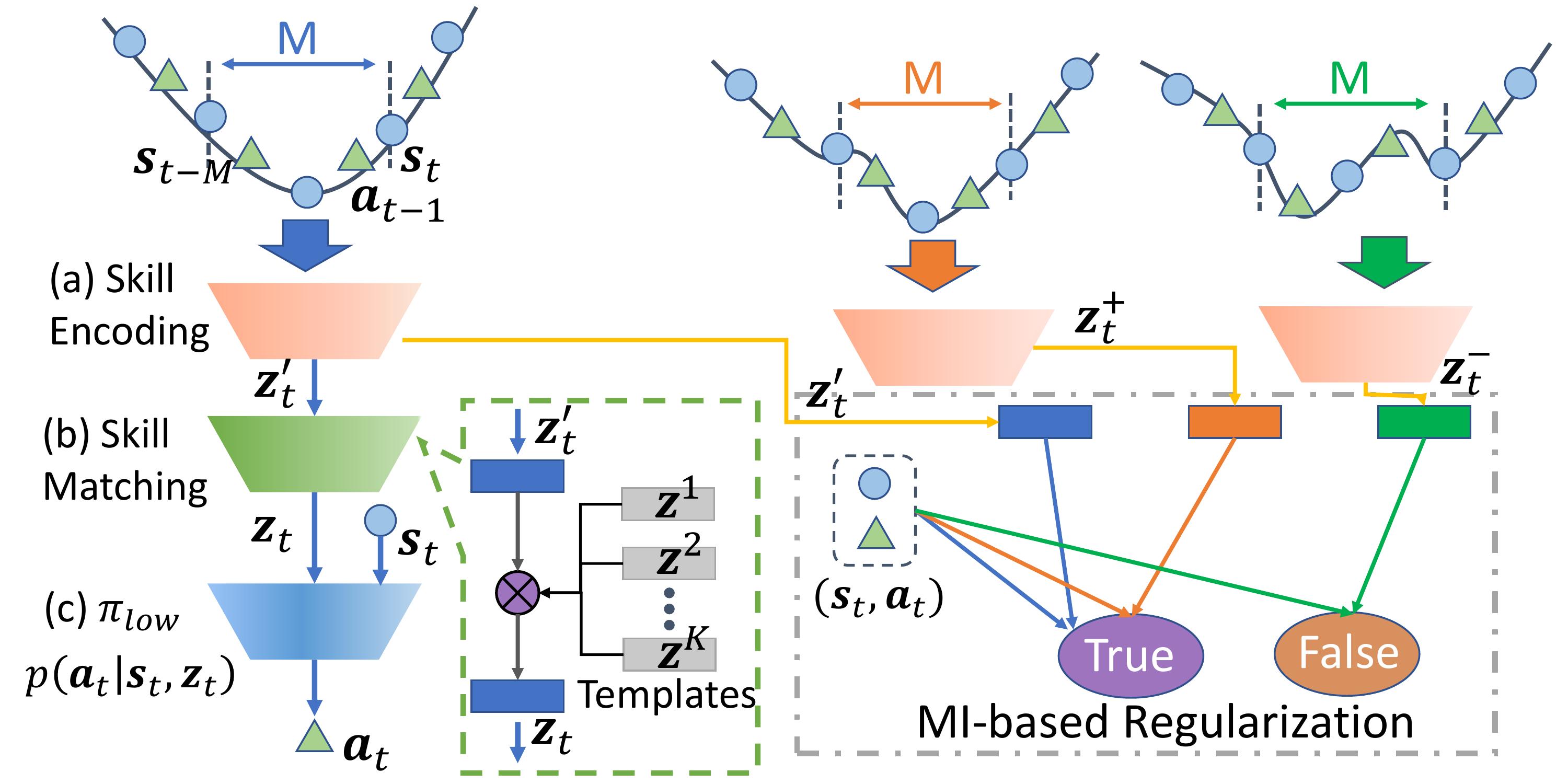}
    \vskip -1em
    \caption{Our framework is composed of three modules: the skill encoding module, skill matching module $g$, and skill-conditioned policy module $\pi_{low}$. The skill encoding module uses previous time steps with window size $M$ as input, and the skill matching module contains $K$ skill templates as parameters. A mutual-information-based regularization is designed to promote automatic skill discovery. A novel skill-based positive-unlabeled (PU) learning along with deep embedding clustering (DEC) is designed to select positive and negative pairs ($\vz^+$ and $\vz^-$) for learning a skill set of varying optimality.  }\label{fig:frame}
    \vskip -1em
\end{figure}

An overview of our hierarchical framework is shown in Figure~\ref{fig:frame}, in which a set of variables $\{\vz^k \in \R^{d_z}, k=[1,\dots,K] \}$ is used to parameterize skills. $d_z$ is the dimension of skill embeddings, and $K$ is the total number of skill variables. The inference of {\method} follows two steps: (1) a high-level policy $\pi_{high}(\vz_t \mid \dots, \vs_{t-1},\va_{t-1}, \vs_{t})$ that selects the skill $\vz_t$ for time step $t$ based on the historical transitions, corresponding to the skill encoding and skill matching modules in Fig.~\ref{fig:frame}; (2) a low-level skill-conditioned policy module $\pi_{low}(\va_t \mid \vs_{t}, \vz_{t})$ which predicts the to-be-taken actions, as in Fig.~\ref{fig:frame}. %Next, we go into details of this framework design.
Concretely, details of these three modules are introduced below:
\begin{itemize}[leftmargin=0.1in]
    \item  A skill encoding module that maps historical transitions and current states to the skill embedding space $\R^{d_z}$. We use $\vs_t$ and a sequence of state-action pairs $[\vs_{t-M}, \va_{t-M}, \dots, \vs_{t-1}, \va_{t-1}]$ as the input to obtain the latent skill embedding $\vz_{t}'$. $M$ is the length of the look-back window. To enable quick skill discovery in account of transition dynamics, we further incorporate states of the next step $\vs_{t+1}$ as an auxiliary input during the skill discovery phase following~\cite{hakhamaneshi2021hierarchical}, and the encoder can be modeled as $p_{f_{bi}}(\vz_t' \mid \vs_{t-M}, \va_{t-M}, \dots, \vs_t, \vs_{t+1})$. In the skill reusing phase, as future states should not available, we model the encoder as $p_{f_{uni}}(\vz_t' \mid \vs_{t-M}, \va_{t-M}, \dots, \vs_t)$.
    \item A skill matching module $g$ that maintains a set of K prototypical embeddings $\{\vz^1, \vz^2, \dots, \vz^K\}$ as $K$ skills. In the inference of time step $t$, extracted skill embedding $\vz_t'$ would be compared with these prototypes and get mapped to one of them to generate $\vz_t$, with the distribution probability following: 
    \begin{equation}\label{eq:skill_sel}
        p(\vz_t=\vz^k) = \frac{1/D(\vz_t', \vz^k)}{\sum_{i=1}^K1/D(\vz_t',\vz^i)}.
    \end{equation} 
    In this equation, $D(\cdot)$ is a distance measurement in the skill embedding space and we stick to the Euclidean distance in this work. To encourage the separation of skills and increase its interpretability, we use hard selection in the generation of $\vz_t$. However, gradients are not readily available for the selection operation. Addressing this problem, we adopt Gumbel softmax~\cite{jang2017categorical}, in which the index of selected $\vz$ is obtained following: 
    \begin{equation}\label{eq:gumbel}
    \begin{aligned}
        index_{\vz} &= \argmax_i \frac{\exp((G_i + \log(p(\vz_t=\vz^i)))/t)}{\sum_j \exp((G_j+\log(p(\vz_t=\vz^j)))/t)}, \\
        G_i & \sim Gumbel(0,1).
    \end{aligned}
    \end{equation}
     $G_i$ is sampled from the standard Gumbel distribution and $t$ here represents temperature typically set to $1$. The re-parameterization trick in ~\cite{jang2017categorical} enables differentiable inference of Eq.~\ref{eq:gumbel}, which allows these prototypical skill embeddings to be updated along with other parameters in the learning process.
    \item A low-level policy module $\pi_{low}$ that captures the mapping from state to actions conditioned on the latent skill variable. It takes the state $\vs_t$ (current observation) and skill variable $\vz_t$ (skill to use) as the input, and predicts the action: $p_{\pi_{low}}(\va_t \mid \vs_t, \vz_t)$. The imitation learning loss can be computed as:
\begin{equation}\label{eq:imi_loss}
    \mathcal{L}_{imi} = -\E_{\tau_i}\E_{(\vs_t,\va_t)\in \tau_i}\E_{\vz_t \sim \pi_{high}} \log p_{\pi_{low}}(\va_t \mid \vs_t, \vz_t),
\end{equation}
which takes a hierarchical structure and maximizes action prediction accuracy on given demonstrations.
\end{itemize}

The high-level policy $\pi_{high}$ is modeled by bi-directional skill encoding module $f_{bi}$ and skill matching module $g$ in the first phase, and by uni-directional skill encoding module $f_{uni}$ and $g$ in the second phase. We will go into detail about learning of these two phases in the following sections. 

\subsection{Discovering the Skill Set }
In the skill discovery phase, we target to imitate demonstrations of $\mathcal{D}^{clean} \cup \mathcal{D}^{noisy}$ with the designed hierarchical framework, modeling the dynamics in action-taking strategies with explicit skill variables. However, directly using the imitation loss in Eq.~\ref{eq:imi_loss} alone is insufficient to learn a skill set of varying optimality. There are some further challenges:
\begin{itemize}[leftmargin=0.2in]
    \item Each skill variable $\vz^k$ may degrade to modeling an average of the global policy, instead of capturing distinct action-taking strategies from each other~\cite{celik2022specializing}. A sub-optimal high-level policy $\pi_{high}$ could tend to select only a small subset of skills or query the same skill for very different states.
    \item As collected transitions are of varying qualities, the extracted skill set is expected to contain both high-quality skills and low-quality skills. However, ground-truth optimality scores of transitions from $\mathcal{D}^{noisy}$ are unavailable, rendering extra challenges in differentiating and evaluating these skills. 
\end{itemize}

To address these challenges, we design a set of countermeasures. First, to encourage the discovery of specialized skills that are distinct from each other, we design a mutual-information-based regularization term. Second, to guide the skill selection and estimate segment optimality, we further refine the skill discovery process by deep clustering and skill optimality estimation with Positive-Unlabeled (PU) learning. In the end, we incorporate $\vs_{t+1}$ during skill encoding to take the inverse skill dynamics into consideration. Next, we will go into their details in this section.

\subsubsection{Mutual-Information-based Regularization} 
To encourage the discovery of distinct skills, we propose a mutual information (MI) based regularization in the skill discovery phase. Each skill variable $\vz^k$ should encode a particular action policy, corresponding to the joint distribution of states and actions $p(\vs, \va \mid \vz^k)$. From this observation, we propose to maximize the mutual information between the skill $\vz$ and the state-action pair $\{\vs, \va\}$: $\max I((\vs,\va),\vz)$. Mutual information (MI) measures the mutual dependence between two variables from the information theory perspective, and formally it can be written as:
%\begin{equation}\label{eq:MI}
%    I(X, Y) = \int_{\sX \times \sY} log\frac{dP_{XY}}{dP_X \otimes P_Y} dP_{XY},
%\end{equation}
\begin{equation}\label{eq:MI}
\begin{aligned}
    I((\vs,\va), \vz) = & \int_{\mathcal{S}\times\mathcal{A}}\int_{\mathcal{Z}} p(\vs,\va,\vz) \\
    & \cdot \log\frac{p(\vs,\va,\vz)}{p(\vs,\va) \cdot p(\vz)} d(\vs,\va) d\vz,
\end{aligned}
\end{equation}
in which $p(\vs,\va,\vz)$ is the joint distribution probability, and $p(\vs,\va)$ and $p(\vz)$ are the marginals. This mutual information objective can quantify how much can be known about $(\vs, \va)$ given $\vz$, or symmetrically, how much can be known about $\vz$ given the transition $(\vs, \va)$. Maximizing this objective corresponds to encouraging each skill variable to encode an action-taking strategy that is identifiable, and maximizing the diversity of the learned skill set.

MI can not be readily computed for high-dimension data due to the probability estimation and integration in Eq.~\ref{eq:MI}. Following~\cite{belghazi2018mutual,hjelm2018learning,zhao2023towards}, we design a JSD-based MI estimator and formulate the regularization term as follows: 
\begin{equation}
    \begin{aligned}\label{eq:Lmi}
    \mathcal{L}_{mi} =& \E_{\tau_i}\E_{(\vs_t,\va_t)\in \tau_i}[\E_{\vz_i^+} sp(-T(\vs_t, \va_t, \vz_i^+)) \\
    &+ \E_{\vz_i^-} sp(T(\vs_t, \va_t, \vz_i^-))].
    \end{aligned}
\end{equation}
In this equation, $T(\cdot)$ is a compatibility estimation function implemented as a MLP, and $sp(\cdot)$ is the \textit{softplus} activation function. $\vz_i^+$ represents the skill selected by $(\vs_i,\va_i)$ that is a \textbf{positive pair} of $(\vs_t,\va_t)$. On the contrary, $\vz_i^-$ denotes the skill selected by $(\vs_i,\va_i)$ that is a \textbf{negative pair} of $(\vs_t,\va_t)$. A positive pair denotes the transition that is similar to $(\vs_t,\va_t)$ in both embedding and optimality quality, and contrarily is the negative pair. Details of obtaining these positive or negative pairs will be provided in Sec~\ref{sec:DEC&PU}. This regularization can encourage different skill variables to encode different action policies, and those positive pairs should select similar skills while negative pairs should select different skills.

\subsubsection{Positive and Negative Pair Discovery}~\label{sec:DEC&PU}
The optimization of mutual information regularization in Eq.~\ref{eq:Lmi} requires obtaining positive and negative pairs to learn a diverse skill set. One direct solution would be using itself as the positive pair (use $\vz_t$ in place of $\vz_i^+$ in Eq.~\ref{eq:Lmi}) and the skills used by randomly sampled other transitions as the negative pair, as conducted in~\cite{hjelm2018learning}. However, such strategy neglects potential guiding information and may selects transitions using the same skill as negative pairs, introducing noises into the learning process. To promote the automatic skill discovery, instead of randomly sampling we propose to utilize two heuristics: similarity and estimated optimality of transitions.

\begin{figure}
    \centering
    \includegraphics[width=\linewidth]{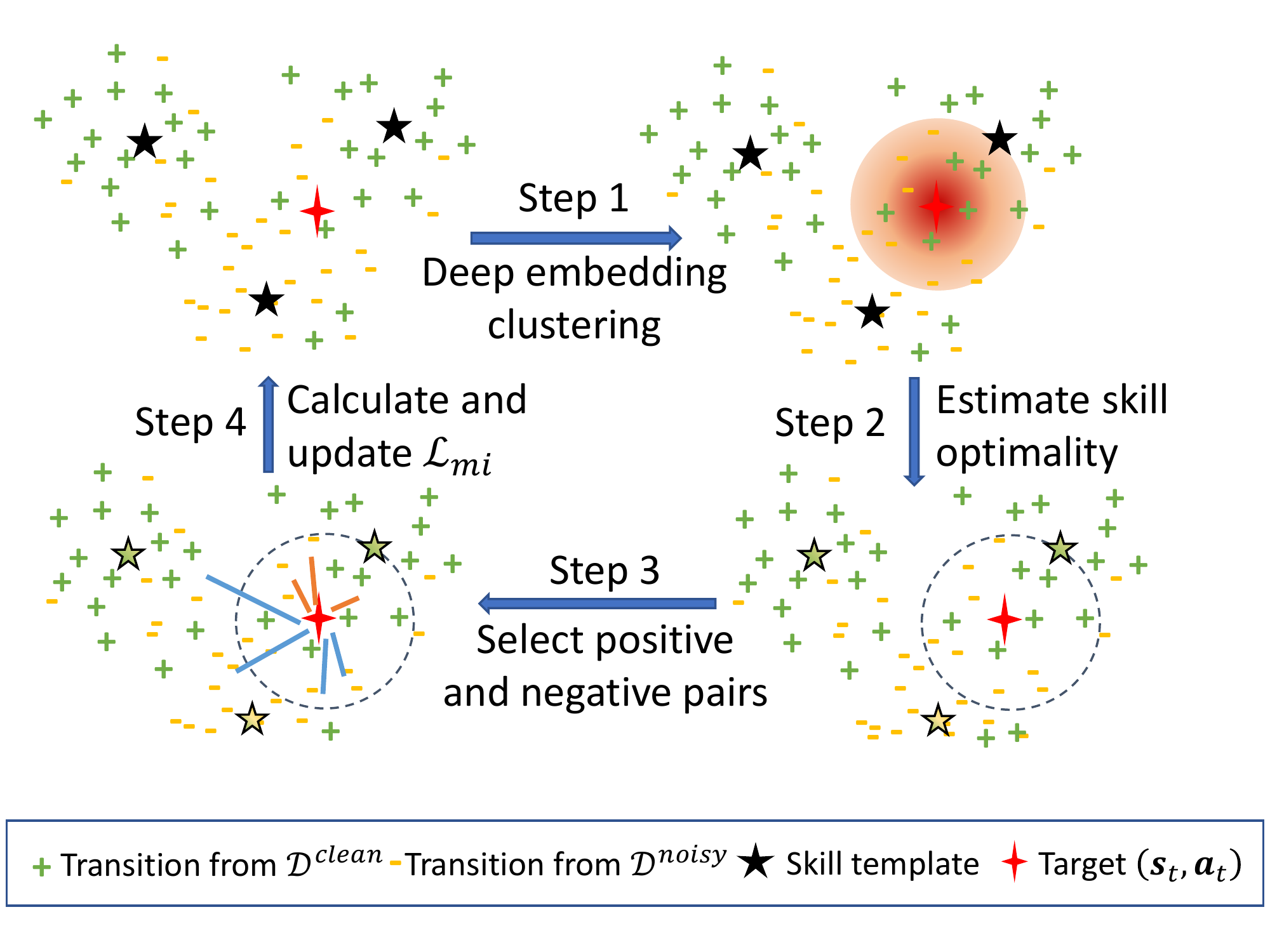}
    \vskip -1.5em
    \caption{A high-level illustration of our strategy in selecting positive and negative pairs for $(\vs_t,\va_t)$. In the first step, DEC is used to find transitions similar to the target (shown with a heatmap). Then, the optimality of skills will be estimated with PU learning (shown with different colors). Following that, estimated optimality scores would be propagated to transitions of $\mathcal{D}^{noisy}$, and our selection strategy is designed based on them (orange edges for positive pairs and blue edges for negative pairs). }
    \label{fig:DEC_PU}
    \vskip -1em
\end{figure}

Concretely, we design a dynamic approach for identifying positive and negative pairs based on those two heuristics. The proposed strategy is composed of two parts, (1) a deep clustering (DEC) algorithm that can discover latent groups of transitions and capture their similarities, which will encourage different skill variables to encode the action primitives of different transition groups; (2) a PU learning algorithm that utilizes both $\mathcal{D}^{clean}$ and $\mathcal{D}^{noisy}$ to evaluate the optimality of discovered skills, and can propagate estimated optimality scores to transitions. An illustration is provided in Fig.~\ref{fig:DEC_PU}. Next, we will go into detail about these two algorithms.

To find similar transitions, we first propose to take a deep embedding cluster (\textbf{DEC}) strategy, by measuring the distance in the high-dimensional space extracted by the skill encoding module $f_{bi}$. Denoting the distance between $(\vs_t, \va_t)$ and $(\vs_i, \va_i)$ as $D(\vz_t',\vz_i')$, we obtain the candidate positive group for $\vz_t$ as the those transitions with a small distance from it, and candidate negative group as those transitions with a large distance. This design will encourage those transitions taken similarly by the skill encoding module to select similar skills, and contrarily for dissimilar ones. Note that measured distances in the embedding space could be very noisy at the beginning, and their quality would be gradually improved during training. Therefore, we add a proxy approach by applying the clustering algorithm directly to the input states, and use variable $\zeta$ to control the probability of adopting DEC or this pre-computed version. In training, we will gradually increase $\zeta$ to shift from this pre-computed clustering to DEC.

Next, we propose to obtain pseudo optimality scores and refine the candidate positive pairs with a Positive-Unlabeled (\textbf{PU}) learning scheme~\cite{bekker2020learning}. As $\mathcal{D}^{noisy}$ contains sub-optimal demonstrations with transitions taking imperfect actions, it is important to differentiate transitions of varying qualities and imitate them with different skills. However, one major challenge is that ground-truth evaluations of those transitions are unavailable. We only have transitions from $\mathcal{D}^{clean}$ as positive examples and transitions from $\mathcal{D}^{noisy}$ as unlabeled examples. Addressing it, we design a skill-based PU learning algorithm. Optimality scores of discovered skills will first be estimated and then be propagated to those unlabeled transitions. 

Concretely, we first estimate the optimality score of skills based on the preference of expert demonstrations and the action prediction accuracy. Intuitively, those skills preferred by expert demonstrations over noisy demonstrations and have a high action prediction accuracy would be of higher quality%\suhang{this seems to be problematic. Consider the patient trajectory. Intuitively, depending on the state of the patient, we would select different skills. However, most patient don't have severe symptoms, and thus can be cured with skill A. But it doesn't mean that skill A is the optimal for every patient}. 
Then we propagate these scores to unlabeled transitions based on their skill selection distributions. The estimated optimality score will also evolve with the training process. The detailed algorithm steps are as follows:
\begin{itemize}[leftmargin=0.2in]
    \item First, we obtain the skill selection distribution and denote it as $P^{\vz}=\{p^{\vz}_k, k\in [1, \dots, K] \}$. We can calculate the selection distribution of clean demonstrations as:
    \begin{equation}
    p^{\vz,clean}_k = \E_{\tau_i \in \mathcal{D}^{clean}}\E_{(\vs_t,\va_t)\in \tau_i} p(\vz_t=\vz^k),    
    \end{equation} 
    and of noisy demonstrations as:
    \begin{equation}
        p^{\vz,noisy}_k = \E_{\tau_i \in \mathcal{D}^{noisy}}\E_{(\vs_t,\va_t)\in \tau_i} p(\vz_t=\vz^k).
    \end{equation} 
    \item Then, the expert preference score $s^{pref}_k$ of skill $k$ can be computed as $(p^{\vz,clean}_k-p^{\vz,noisy}_k)/(p^{\vz,clean}_k+\delta)$, in which $\delta$ is a small constant to prevent division by $0$. 
    \item The quality score of each skill can be computed based on its action-prediction accuracy when selected: 
    \begin{equation}
        s^{qual}_k = \E_{\tau_i}\E_{(\vs_t,\va_t)\in \tau_i}\E_{\vz_t=\vz^k}p(\va_t \mid \vs_t, \vz^k).
    \end{equation}
     
    \item The estimated optimality score $s^{op}_k$ of skill $k$  can be calculated by normalizing the product of these two scores $s^{pref}_k \cdot s^{qual}_k$ into the range $[-1,1]$.
    \item With discovered skills evaluated, optimality scores will be propagated to each transition of $\mathcal{D}^{noisy}$, based on the skill it selected and its performance. For transition $(\vs_t,\va_t)$, we compute its optimality as: $\sum_{k=1}^{K}p(\vz_t=\vz^k)\cdot s^{op}_k$. 
    
\end{itemize}
All transitions of $\mathcal{D}^{clean}$ have optimality score set to $1$. We would refine the candidate positive group of $\vz_t$ by removing those that have a very different optimality score with a threshold $\epsilon$. Note that this process is not needed for the candidate negative group, as they should be encouraged to select different skills regardless of optimality. The estimation of skill optimality scores is updated every $N_{PU}$ epochs during training to reduce the instability. The algorithm is provided in Alg.~\ref{alg:MI}.

\subsubsection{Incorporating Next-step Dynamics}
The core target of this work is to discover latent action-taking strategies from collected demonstrations and encode them explicitly. One problem is, as $\mathcal{D}^{noisy}$ is noisy and contains demonstrations of varying qualities, there may exist transitions taking very different actions with similar observations and histories. Due to the lack of ground-truth optimality score, it could be difficult for the skill encoding module to tell them apart and differentiate their latent skills. Therefore, in this skill discovery phase, we also include $\vs_{t+1}$ as input to the skill encoding module, so that skills can be encoded in an influence-aware manner~\cite{hakhamaneshi2021hierarchical,zhao2022exploring}. $\vs_{t+1}$ enables the skill selection to be not only conditioned on the current and prior trajectories, but also on a future state, which can help to differentiate skills that work in similar states. We denote this bi-directional skill encoder as $f_{bi}$, as in Fig.~\ref{fig:frame}. Note that $\vs_{t+1}$ is only used in the skill discovery phase, hence will not result in the information leakage problem.

Putting all together, in the skill discovery phase we will train modules $\{f_{bi}(\cdot), g(\cdot), \pi_{low}(\cdot)\}$ on $\mathcal{D}^{clean} \cup \mathcal{D}^{noisy}$, and use a mutual information loss to encourage the learning of a diverse skill set. Two algorithms are designed to estimate the similarity and optimality of transitions, and are used to refine the computation of the regularization term in Eq.~\ref{eq:Lmi}. The full learning objective is:
\begin{equation}
    \min_{f_{bi},g,\pi_{low}}\max_{T} \mathcal{L}_{imi} + \lambda \cdot \mathcal{L}_{mi},
\end{equation}
where $T$ is the compatibility estimator used in MI estimation (Eq.~\ref{eq:Lmi}). A detailed update step is provided in Alg.~\ref{alg:MI}.

\subsection{Reusing Skills for Expert Policy}
With the skill discovery phase completed, we would utilize the learned skill set to imitate expert demonstrations in $\mathcal{D}^{clean}$. Specifically, we will adapt the modules $\{f_{uni}(\cdot), g(\cdot), \pi_{low}(\cdot) \}$ by imitating $\mathcal{D}^{clean}$ in this phase. Concretely, as $\{f_{bi}, g(\cdot), \pi_{low}(\cdot) \}$ are already learned in the skill discovery phase, we would split this \textit{skill-reusing} into two steps: (1) In the first step, we would freeze the parameters of $\{g(\cdot), \pi_{low}(\cdot) \}$, which contain extracted skills and skill-conditioned policies, and only learn $\{f_{uni}(\cdot) \}$ on $\mathcal{D}^{clean}$ to obtain a high-level skill selection policy. This step utilizes pre-trained skills to mimic expert demonstrations $\mathcal{D}^{clean}$. Besides this imitation loss in Eq.~\ref{eq:imi_loss}, we further propose to transfer the skill selection knowledge from $f_{bi}$ to $f_{uni}$:
\begin{equation}
    \mathcal{L}_{KD} = -\E_{\tau_i}\E_{(\vs_t,\va_t)\in \tau_i}p(\vz_t = \bar{\vz_t}), 
\end{equation}
in which $\bar{\vz_t}$ is predicted using $f_{bi}$. For simplicity, we do not manipulate the weight of $\mathcal{L}_{KD}$ as it has the same scale as $\mathcal{L}_{imi}$, and the learning objective is 
\begin{equation}
\min_{f_{uni},g,\pi_{low}} \mathcal{L}_{imi} + \mathcal{L}_{KD}.
\end{equation}
(2) In the second step, we will further refine the whole framework in an end-to-end manner, based on the imitation objective $\mathcal{L}_{imi}$.

Aside from fine-tuning the skill-based framework on $\mathcal{D}^{clean}$, we further propose to utilize transitions from $\mathcal{D}^{noisy}$ that have a low optimality score~\cite{wang2020adversarial}. In the skill discovery phase, a PU learning algorithm is conducted iteratively to evaluate the quality of transitions from $\mathcal{D}^{noisy}$, and will assign an optimality score to each of them. We can collect transitions with low optimality scores from demonstrations in $\mathcal{D}^{noisy}$ as $\mathcal{D}^{neg}$, and a new optimization objective $\mathcal{L}_{adv}$ can be designed to encourage our agent to perform differently:
\begin{equation}
    \min_{f_{uni},g,\pi_{low}}\mathcal{L}_{adv} = \E_{(\vs_t,\va_t)\in \mathcal{D}^{neg}}\log p(\va_t \mid \vs_t,\vz_t).
\end{equation}
In experiments, we set a hard threshold to collect $\mathcal{D}^{neg}$, and the learning objective becomes $\mathcal{L}_{imi}+\mathcal{L}_{adv}$. This objective can encourage the model to prevent making the same mistakes as those low-quality demonstrations. 

\subsection{Training Algorithm}
%Putting everything together, we provide the detailed algorithm for training our framework {\method} in Alg.~\ref{alg:Framwork}. Due to limitation of space, we put it in Appendix.~\ref{ap:alg}. \suhang{seems that you have enough space. bring these the algorithms back, but shorten the algorithms a little bit?}
\begin{algorithm}[t]
  \caption{Skill-based IL from Imperfect Demonstrations %\suhang{can you use algorithm format without end if, end for to save space.}
  }
  \label{alg:Framwork}
  %\scalebox{0.75}{
  \begin{algorithmic}[1] 
  \REQUIRE %??????????Input
    Expert demonstrations $\mathcal{D}^{clean}$, Noisy demonstrations $\mathcal{D}^{noisy}$, %Randomly initialized skill encoding module $f_{bi}$ and $f_{uni}$, Randomly initialized skill matching module $g$ with random initialization on latent skill set $\{\vz^1, \vz^2, \dots, \vz^K\}$, Randomly initialized skill-conditioned policy $\pi_{low}$, 
    Pre-train epochs $N$, PU interval $N_{PU}$
    \STATE Random initialize model parameters
    \FOR{$t$ in $N$}
    \STATE Optimize $f_{bi}, g$ and $\pi_{low}$ to discover skills following Alg.~\ref{alg:MI}%, with:\\ $\mathcal{L}_{imi} + \lambda \cdot \mathcal{L}_{mi}$ 
    \IF{$t\%N_{PU}=0$}
    \STATE Update optimality estimation of skills, as Sec.~\ref{sec:DEC&PU}
    \ENDIF
    \ENDFOR
    \WHILE{Not Converged}
    \STATE Freeze parameters in $g(\cdot)$ and $\pi_{low}(\cdot)$
    \STATE Learn $f_{uni}$ with $\mathcal{L}_{imi} + \mathcal{L}_{KD}$ on $\mathcal{D}^{clean}$
    \ENDWHILE
    \WHILE{Not Converged}
    \STATE Update $f_{uni}, g$ and $\pi_{low}$ with $\mathcal{L}_{imi}+\mathcal{L}_{adv}$ 
    \ENDWHILE
    \RETURN Trained $f_{uni}$, $g$ and $\pi_{low}$.
  \end{algorithmic}
  %}%% resizebox
\end{algorithm}

The training algorithm is provided in Alg.~\ref{alg:Framwork} and Alg.~\ref{alg:MI}. The skill discovery phase corresponds to line $2-7$ in  Alg.~\ref{alg:Framwork}, and the details of learning with MI-based regularization in line $3$ is presented in Alg.~\ref{alg:MI}. Mi-based regularization will help {\method} to learn a set of disentangled skills. Then, in the skill reusing phase, we first freeze the learned skills to update the high-level policy in line $8-11$, and then we finetune the whole framework end-to-end in line $12-14$. %\suhang{I changed Algorithm 1, please revise the line number accordingly} \suhang{add a little bit description.   Could you prepare a version by shortening Algorithm 2 and combine algorithm 1 with 2? Then let's choose which version to use.}

\begin{algorithm}[t]
  \caption{MI-augmented Skill Discovery Step}
  \label{alg:MI}
  %\scalebox{0.75}{
  \begin{algorithmic}[1] 
  \REQUIRE 
  Transition set $\{(\vs_t, \va_t)\}$ from $\mathcal{D}^{clean}$ and $\mathcal{D}^{noisy}$, skill encoding module $f_{bi}$, skill matching module g, skill-conditioned policy model $\pi_{low}$, MI compatibility estimation function $T$
  \STATE Draw b transition samples  %the given data distribution: 
  $\{(\vs_t, \va_t)\}_{t=1}^b \sim \mathcal{D}^{clean} \cup \mathcal{D}^{noisy}$
  \STATE For each $(\vs_t,\va_t)$, sample its candidate positive pairs $(\vs_t^{+},\va_t^+)$ from the same clustering group
  \STATE Filter candidate positive pairs based on estimated optimality score following Sec.~\ref{sec:DEC&PU}
  \STATE For each $(\vs_t,\va_t)$, sample its negative pairs $(\vs_t^{-},\va_t^-)$ of different clustering groups
  \STATE Estimate the mutual information loss $\mathcal{L}_{mi}$ following Eq.~\ref{eq:Lmi}%: \\ $\E_{(\vs_t,\va_t)}[\E_{\vz_i^+} sp(-T(\vs_t, \va_t, \vz_i^+)) + \E_{\vz_i^-} sp(T(\vs_t, \va_t, \vz_i^-))]$
  \STATE Update compatibility function $T$ as %to maximize $\mathcal{L}_{mi}$: \\
  $T \leftarrow T+ \nabla_{T} \mathcal{L}_{mi}$
  \STATE Update $f_{bi}$, $g$ and $\pi_{low}$ by: 
  $\min_{f_{bi},g,\pi_{low}} \mathcal{L}_{imi} + \lambda \cdot \mathcal{L}_{mi} $
  \RETURN Updated $f_{bi}$, $g$ and $\pi_{low}$, $T$
  \end{algorithmic}
  %}%% resizebox
\end{algorithm}

%% file: secs/experiment.tex
\section{Experiment}
    In this section, we conduct experiments to evaluate the proposed imitation learning framework in discovering skills and learning high-quality policies from sub-optimal demonstrations. Specifically, in these experiments, we want to answer the following questions:
    \begin{itemize}[leftmargin=0.1in]
        \item \textbf{RQ1}: Can our framework {\method} achieve stronger performance in imitating demonstrations of varying qualities?
        \item \textbf{RQ2}: Is the proposed skill discovery algorithm able to disentangle latent skills of varying optimality apart?
        \item \textbf{RQ3}: Can {\method} provide improved interpretability on the knowledge learned from demonstrations?
    \end{itemize}

\subsection{Experiment Settings}
    \subsubsection{Datasets}
    We evaluate the proposed framework on three datasets, MiniGrid-FourRoom, DoorKey~\cite{gym_minigrid}, along with a public EHRs dataset Mimic-IV~\cite{bajor2016predicting}. In FourRoom and Doorkey, the agent needs to navigate in a maze and arrive at the target position to receive a reward. In Mimic-IV, each demonstration is a trajectory of medication actions received by a patient, with physiological features as states and treatments as actions. Details of these datasets are provided in Appendix.~\ref{ap:datasets}.

\subsubsection{Baselines}
%Our proposed {\method} is compared with representative imitation learning methods and state-of-the-art treatment recommendation methods. %including Behavior Cloning (BC)~\cite{pomerleau1998autonomous}, DensityIRL~\cite{wang2020imitation}, MCE IRL~\cite{ziebart2010modeling}, AIRL~\cite{fu2018learning} and GAIL~\cite{ho2016generative}. Furthermore, We also compare {\method} with two recent strategies for utilizing noisy demonstrations explicitly: ACIL~\cite{wang2020adversarial} and DWBC~\cite{xu2022discriminator}. An introduction of these methods is provided in Appendix.~\ref{ap:baseline}.
First, we compare {\method} to representative imitation learning and treatment recommendation methods:
\begin{itemize}[leftmargin=0.1in]
    \item Behavior Cloning (BC)~\cite{pomerleau1998autonomous}: BC bins the demonstrations into transitions and learns an action-taking policy in a supervised manner. 
    %\item SRL-RNN~\cite{wang2018supervised}: It manually designs a sparse reward function and combines supervised learning and reinforcement learning for dynamic treatment recommendations.
    \item DensityIRL~\cite{wang2020imitation}: Density-based reward modeling conducts a non-parametric estimation of marginal as well as joint distribution of states and actions from provided demonstrations, and computes a reward using the log of those probabilities.
    \item MCE IRL~\cite{ziebart2010modeling}: Its full name is Maximum causal entropy inverse reinforcement learning, whose idea is similar to DensityIRL. It designs an entropy-based approach to estimate causally conditioned probabilities for the estimation of reward function.
    \item AIRL~\cite{fu2018learning}: Adversarial inverse reinforcement learning aims to automatically acquiring a scalable reward function from expert demonstrations by adversarial learning.
    %\item D3Q~\cite{raghu2017continuous}: It is similar to the framework of SRL-RNN and trains the policy model with deep Q-learning.
    \item GAIL~\cite{ho2016generative}: GAIL directly trains a discriminator adversarially to provide the reward signal for policy learning.
    %\item Directed Info-Gail (D-GAIL)~\cite{sharma2018directed}: Directed Info-GAIL extends vanilla GAIL by modeling the generation of expert trajectories as of multiple modes in a graphical model and mimics the expert policy by maximizing the directed information flow.
    %\item SHIL~\cite{wang2022hierarchical}: SHIL designs a hierarchical framework with intermediate subgoals to condition the policy, which is expected to capture the switching from the treatment of one symptom to another.
\end{itemize}
The above methods are all proposed assuming demonstrations to be clean and optimal. We further compare {\method} with two recent strategies that are designed to utilize noisy demonstrations explicitly: ACIL~\cite{wang2020adversarial} and DWBC~\cite{xu2022discriminator}. 
\begin{itemize}[leftmargin=0.1in]
    \item ACIL~\cite{wang2020adversarial}: ACIL extends GAIL by incorporating noisy demonstrations as negative examples and learns the policy model to be distinct from them.
    \item DWBC~\cite{xu2022discriminator}:  It trains a discriminator to evaluate the quality of each demonstration trajectory through adversarial training, and conducts a weighted behavior cloning. 
\end{itemize}

Different from them, our method proposes to break each trajectory down into segments and extract useful skills from them hierarchically. For the ablation study, we also implement a variant:
\begin{itemize}[leftmargin=0.1in]
    \item ${\text{\method}}^*$: This variant does not access $\vs_{t+1}$ in the skill discovery, to analyze the importance of incorporating next-step dynamics.
\end{itemize}

\subsubsection{Evaluation Metrics}
We evaluate the trained agent with two strategies, the data-dependent and environment-dependent respectively. Concretely, the adopted metrics are as follows: (1) data-dependent metrics, which are computed on the test set of clean demonstrations. We select accuracy (ACC), macro AUROC, and micro AUROC scores~\cite{bradley1997use}. For the FourRoom dataset, some actions are never taken so we report the Macro F score instead. %Both MacroAUC and MacroF are calculated and averaged per class, hence are more trustworthy in the existence of class imbalance. 
(2) Environment-dependent metrics. For datasets with the interactive environment readily available, we place the trained agent into the environment and report its averaged rewards after $1,000$ rounds of random running.  

\subsubsection{Configurations}
The policy agent is instantiated to have the same network architecture across different methods for a fair comparison. All methods are trained until convergence, with the learning rate initialized to $0.001$. Adam~\cite{kingma2015adam} optimizer is applied for all approaches with weight decay set to $5e$-$4$. Of obtained demonstrations, train:validation:test split is set as $3:2:5$. For the proposed {\method}, the hyper-parameter $\lambda$ which controls the weight of $\mathcal{L}_{mi}$ is set to $1.0$ and the threshold $\epsilon$ of pseudo optimality score from PU learning is set to $0.1$ if not stated otherwise. The look-back window size $M$ is fixed as $5$, and the controlling variable $\zeta$ of DEC is initialized to $0$ and increased by $0.05$ after each training epoch. 

\begin{table*}[t]
  \caption{Results for FourRoom, Mimic-IV and DoorKey. 
  %In setting 1, we utilize only the clean demonstration set. In setting 2, both clean demonstrations $\mathcal{D}^{clean}$ and non-expert demonstrations $\mathcal{D}^{noisy}$ are used. 
  Best performance is highlighted in bold. %Each experiment is randomly ran for $5$ times with both the mean and standard deviation reported.
  } \label{tab:exp_comparative}
    \vskip -1em
  %\tiny
  \centering
  \resizebox{\linewidth}{!}{
  \begin{tabular}{c c ccc cccc c c c}
    \toprule
    %\multicolumn{2}{c}{Part}                   \\
    %\cmidrule(r){1-2}
    & \multirow{2}*{Method}  & \multicolumn{3}{c}{FourRoom} & \multicolumn{4}{c}{Mimic-IV} & \multicolumn{3}{c}{DoorKey} \\
    \cmidrule(lr){3-5}
    \cmidrule(lr){6-9}
    \cmidrule(lr){10-12}
    & ~ & ACC & MacroF & Reward & ACC & MacroAUC & MicroAUC & MacroF & ACC & MacroF & Reward \\
    \midrule
    \multirow{6}{*}{\rotatebox[origin=c]{90}{$\mathbf{\mathcal{D}^{clean}}$ \textbf{Only}} } & BC 
    & $89.7_{\pm 0.26}$ &  $67.2_{\pm 0.18}$ & $27.6_{\pm 0.9}$
    & $43.7_{\pm 0.41}$ & $80.7_{\pm 0.22}$ & $83.3_{\pm 0.18}$ & $16.7_{\pm 0.15}$ & $88.4_{\pm 0.34}$ &  $85.8_{\pm 0.21}$ & $83.4_{\pm 2.1}$\\
    %& SRL-RNN \\
    & DensityIRL & $86.4_{\pm0.27}$ & $66.5_{\pm0.22}$ & $18.2_{\pm1.5}$ & $38.5_{\pm0.28}$ & $77.9_{\pm0.16}$ & $_{80.8\pm0.19}$ & $_{14.2\pm0.14}$ & $81.7_{\pm0.22}$ & $83.7_{\pm0.16}$ & $75.5_{\pm3.4}$ \\
    & MCE IRL & $88.9_{\pm0.26}$ & $67.3_{\pm0.19}$ & $19.8_{\pm1.6}$ & $41.1_{\pm0.21}$ & $79.4_{\pm0.16}$ & $81.9_{\pm0.17}$ & $15.9_{\pm0.17}$  & $84.5_{\pm0.29}$ & $86.2_{\pm0.23}$ & $78.9_{\pm2.5}$ \\
    & AIRL & $87.1_{\pm 0.22}$ &  $65.8_{\pm 0.23}$ & $23.8_{\pm 1.2}$ &
    $41.5_{\pm 0.36}$ & $80.2_{\pm 0.23}$ & $82.8_{\pm 0.18}$ & $15.8_{\pm 0.16}$ & $89.2_{\pm 0.26}$ &  $86.5_{\pm 0.25}$ & $85.7_{\pm 2.7}$ \\
    & GAIL & $84.5_{\pm 0.33}$ &  $63.2_{\pm 0.31}$ & $18.9_{\pm 1.1}$ &
    $39.2_{\pm 0.31}$ & $76.5_{\pm 0.21}$ & $81.3_{\pm 0.19}$ & $14.6_{\pm 0.18}$ & $80.6_{\pm 0.29}$ &  $81.3_{\pm 0.27}$ & $73.3_{\pm 3.2}$ \\
    %& D-GAIL \\
    %& SHIL \\
    & \method & $91.2_{\pm 0.21}$ &  $68.1_{\pm 0.16}$ & $28.2_{\pm 1.2}$ & $44.6_{\pm 0.32}$ & $81.3_{\pm 0.26}$ & $83.6_{\pm 0.21}$ & $17.2_{\pm 0.18}$ & $89.6_{\pm 0.24}$ &  $87.3_{\pm 0.22}$ & $86.1_{\pm 2.3}$  \\
    \midrule
    \multirow{9}{*}{\rotatebox[origin=c]{90}{$\mathcal{D}_{clean}\cup\mathcal{D}_{noisy}$ }} & BC 
    & $87.2_{\pm 0.24}$ &  $66.4_{\pm 0.15}$ & $26.7_{\pm 1.0}$ 
    &  $41.1_{\pm 0.36}$ & $81.2_{\pm 0.19}$ & $82.9_{\pm 0.16}$ & $16.5_{\pm 0.16}$  & $86.7_{\pm 0.24}$ &  $83.6_{\pm 0.19}$ & $81.8_{\pm 2.3}$  \\
    %& SRL-RNN \\
    & DensityIRL & $85.4_{\pm0.27}$ & $65.4_{\pm0.26}$ & $16.5_{\pm1.3}$ & $36.7_{\pm0.29}$ & $77.6_{\pm0.17}$ & $79.4_{\pm0.21}$ & $13.7_{\pm0.15}$ & $76.8_{\pm0.25}$ & $75.1_{\pm0.19}$ & $71.6_{\pm2.6}$  \\
    & MCE IRL & $87.4_{\pm0.25}$ & $66.3_{\pm0.21}$ & $18.7_{\pm1.5}$ & $39.5_{\pm0.27}$ & $78.9_{\pm0.18}$ & $81.2_{\pm0.19}$ & $14.5_{\pm0.14}$ & $81.3_{\pm0.23}$ & $81.4_{\pm0.23}$ & $74.8_{\pm2.2}$  \\
    & AIRL & $88.7_{\pm 0.28}$ &  $66.7_{\pm 0.33}$ & $25.4_{\pm 1.2}$ & 
    $40.7_{\pm 0.33}$ & $80.4_{\pm 0.23}$ & $82.1_{\pm 0.17}$ & $15.7_{\pm 0.16}$ & $87.5_{\pm 0.31}$ &  $84.9_{\pm 0.28}$ & $82.1_{\pm 2.2}$ \\
    & GAIL & $83.4_{\pm 0.29}$ &  $62.9_{\pm 0.26}$ & $18.8_{\pm 1.1}$ &
    $38.1_{\pm 0.29}$ & $74.8_{\pm 0.22}$ & $80.1_{\pm 0.15}$ & $13.9_{\pm 0.17}$ & $71.7_{\pm 0.22}$ &  $70.8_{\pm 0.17}$ & $70.3_{\pm 2.4}$ \\
    & ACIL & $90.2_{\pm 0.23}$ &  $67.6_{\pm 0.19}$ & $27.5_{\pm 1.1}$
    & $44.2_{\pm 0.35}$ & $81.2_{\pm 0.24}$ & $83.7_{\pm 0.18}$ & $17.1_{\pm 0.15}$ & $89.2_{\pm 0.30}$ &  $87.3_{\pm 0.22}$ & $85.9_{\pm 2.8}$ \\
    %& D-GAIL \\
    %& SHIL \\
    & DWBC & $91.1_{\pm 0.17}$ &  $67.9_{\pm 0.22}$ & $28.3_{\pm 2.1}$ &
    $45.1_{\pm0.37}$ & $81.4_{\pm0.25}$ & $84.3_{\pm0.17}$  & $17.2_{\pm0.16}$  & $90.3_{\pm 0.27}$ &  $88.5_{\pm 0.26}$ & $87.1_{\pm 3.3}$ \\
    & ${\text{\method}}^*$ & ${93.4}_{\pm 0.18}$ &  ${69.1}_{\pm 0.12}$ & ${31.2}_{\pm 1.1}$ & ${48.1}_{\pm 0.31}$ & ${83.8}_{\pm 0.21}$ & ${86.5}_{\pm 0.13}$ & ${20.7}_{\pm 0.16}$  & ${92.9}_{\pm 0.16}$ &  ${91.4}_{\pm 0.25}$ & ${92.3}_{\pm 2.0}$ \\
    & {\method} & $\bf{94.4}_{\pm 0.16}$ &  $\bf{69.5}_{\pm 0.11}$ & $\bf{33.5}_{\pm 1.2}$ & $\bf{49.4}_{\pm 0.26}$ & $\bf{84.4}_{\pm 0.17}$ & $\bf{87.6}_{\pm 0.12}$ & $\bf{22.3}_{\pm 0.14}$ & $\bf{94.6}_{\pm 0.21}$ &  $\bf{92.6}_{\pm 0.18}$ & $\bf{93.2}_{\pm 2.2}$ \\
    
    \bottomrule
  \end{tabular}}
  \vskip -1em
\end{table*}

\subsection{Imitation Learning Performance}
First, we report the performance for action prediction and make the comparison in Table~\ref{tab:exp_comparative}. to answer RQ1. Each experiment is run for $5$ times with random initializations, and we report both the mean and variance w.r.t each metric. Two settings are adopted:
\begin{itemize}[leftmargin=0.1in]
    \item $\mathcal{D}^{clean}$ Only: in the first setting, we test all approaches only on the expert demonstration set, $\mathcal{D}^{clean}$, without utilizing those noisy demonstrations. In this setting, all demonstrations can be trusted to be well-performing.
    \item $\mathcal{D}^{clean} \bigcup \mathcal{D}^{noisy}$: In the second setting, we use the mixed dataset containing both $\mathcal{D}^{clean}$ and $\mathcal{D}^{noisy}$ to augment the training set. This setting can verify whether an algorithm can effectively leverage non-expert demonstrations.
\end{itemize}
It can be observed that our proposed {\method} outperforms all baseline algorithms, especially in the second setting, showing its strong performance in effectively utilizing the noisy demonstrations in addition to the expert ones. For both datasets, incorporating $\mathcal{D}^{noisy}$ would impair the performance of most baseline algorithms with a clear performance gap that can be observed. Such phenomenon suggests that directly imitating $\mathcal{D}^{noisy}$ is undesired and may be misled by noisy demonstrative trajectories. ACIL neglects useful information from $\mathcal{D}^{noisy}$ and takes all of them as negative, hence only achieving marginal improvements. DWBC adopts adversarial training to evaluate trajectory quality and identifies high-quality demonstrations, however it takes each trajectory as a whole and shows only marginal improvement. In the contrast, our method {\method} is able to make better use of $\mathcal{D}^{noisy}$, and shows stronger improvements across various metrics on both datasets. Furthermore, the comparison with variant ${\text{\method}}^*$ validates the effectiveness of incorporating next-step dynamics as auxiliary evidence for the discovery of skills, showing constant improvements across all datasets.

\subsection{Ablation Study}

\begin{figure}
    \centering
    \subfigure[FourRoom, Reward]{\includegraphics[width=0.48\linewidth]{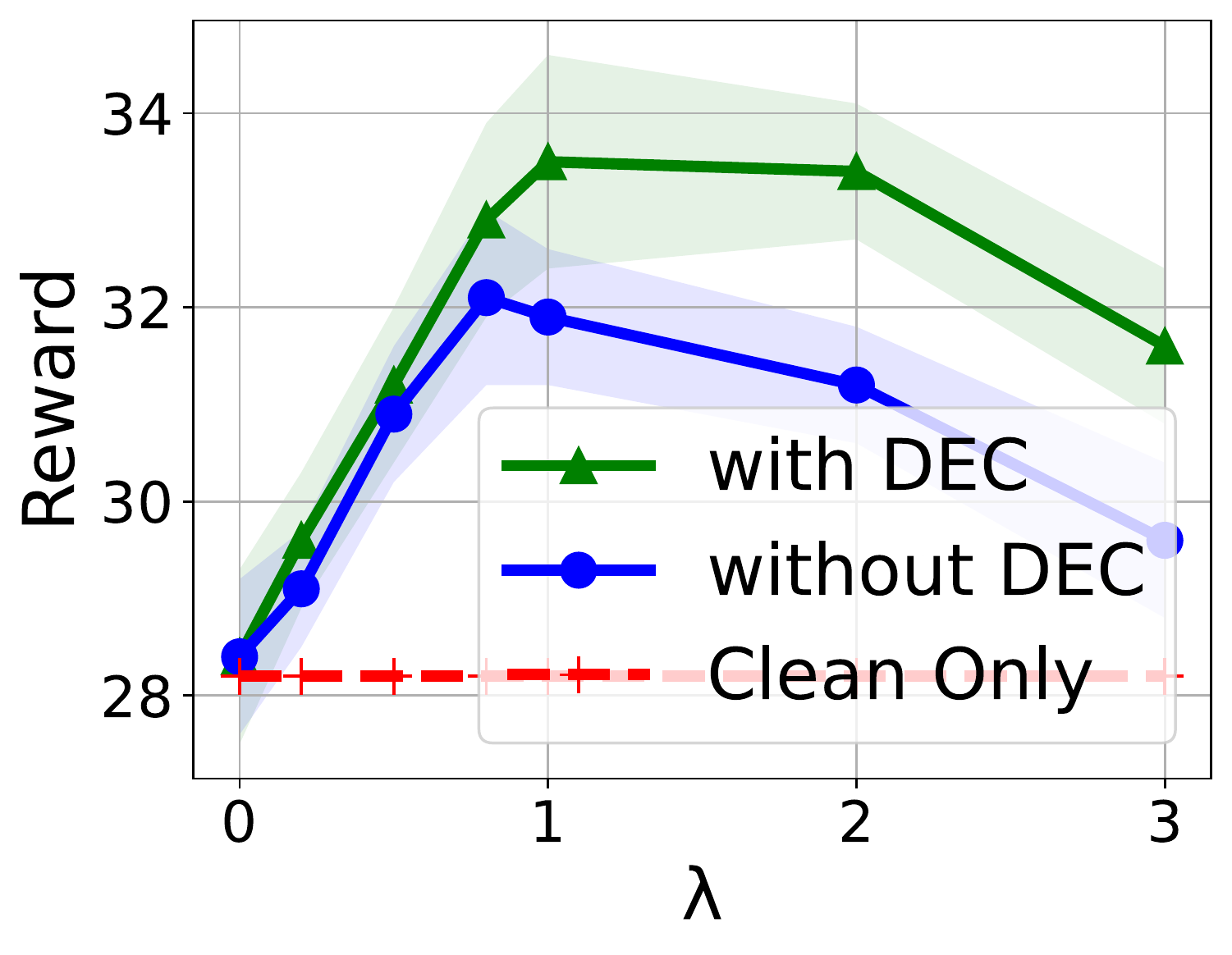}}
    %\subfigure[FourRoom, macroF]{\includegraphics[width=0.32\linewidth]{figs/lambdafourroom_setting0.pdf}}
    \subfigure[Mimic-IV, macroAUC]{\includegraphics[width=0.48\linewidth]{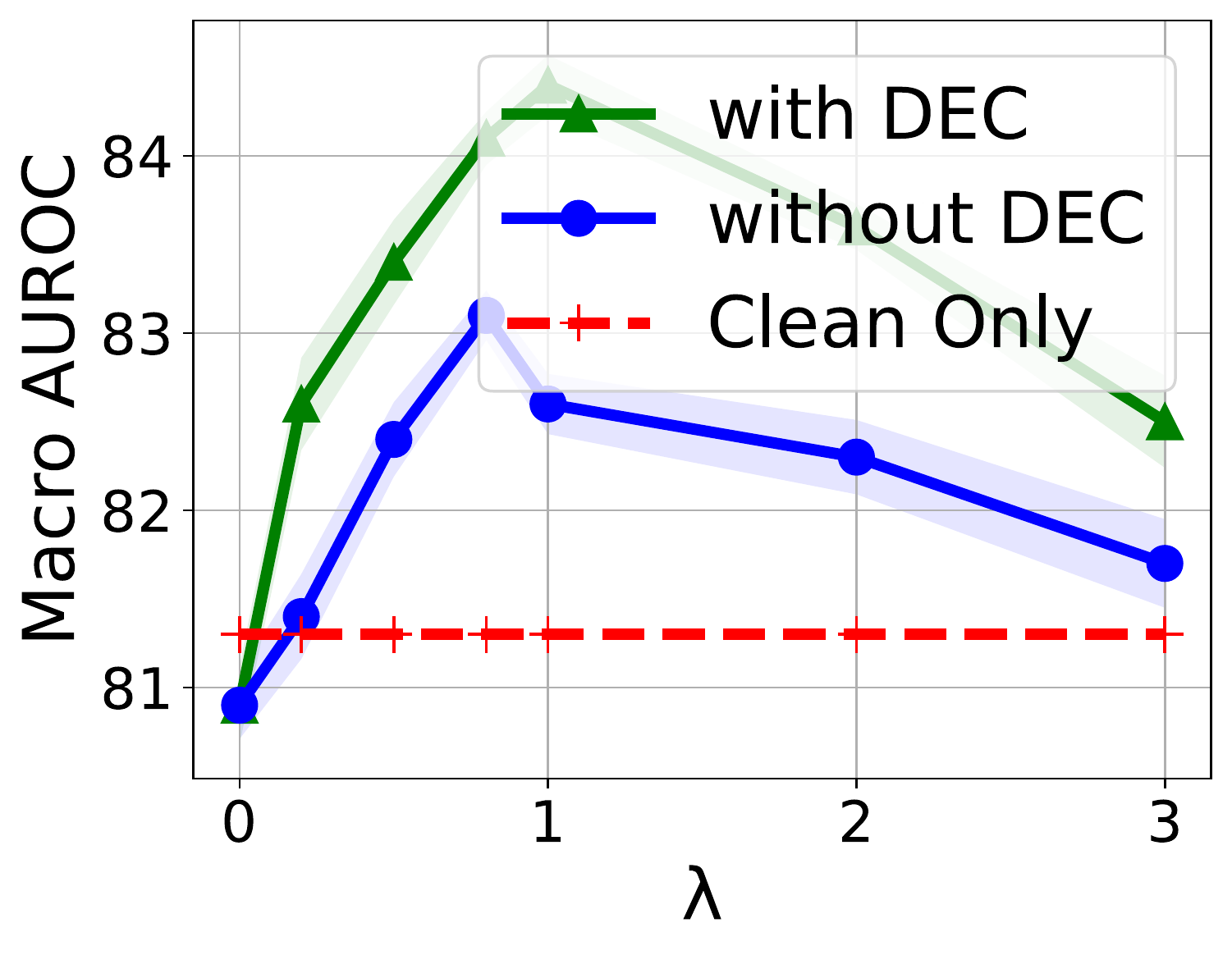}}
    \vskip -1.5em
    \caption{Sensitivity analysis on mutual information regularization $\mathcal{L}_{mi}$ by varying its weight $\lambda$.  }
    \label{fig:mi_sensitivity}
    \vskip -1em
\end{figure}

\textbf{Skill Discovery With Mutual Information.} In {\method}, an MI-based regularization is adopted to encourage the discovery of concrete skills. To evaluate the importance of designed mutual information regularization and partially answer RQ2, we conduct a sensitivity analysis on $\lambda$, which controls the weight of $\mathcal{L}_{mi}$ in the skill discovery phase. Concretely, we vary $\lambda$ in $\{0, 0.2, 0.5, 0.8, 1, 2,3\}$ and leave other hyper-parameters unchanged. Results on both the FourRoom dataset and Mimic-IV are reported in Fig.~\ref{fig:mi_sensitivity}. As a comparison, we show the result w/o DEC part, and also present the performance of the base method trained only with $\mathcal{D}^{clean}$.

\begin{figure}
    \centering
    \subfigure[FourRoom, Reward]{\includegraphics[width=0.48\linewidth]{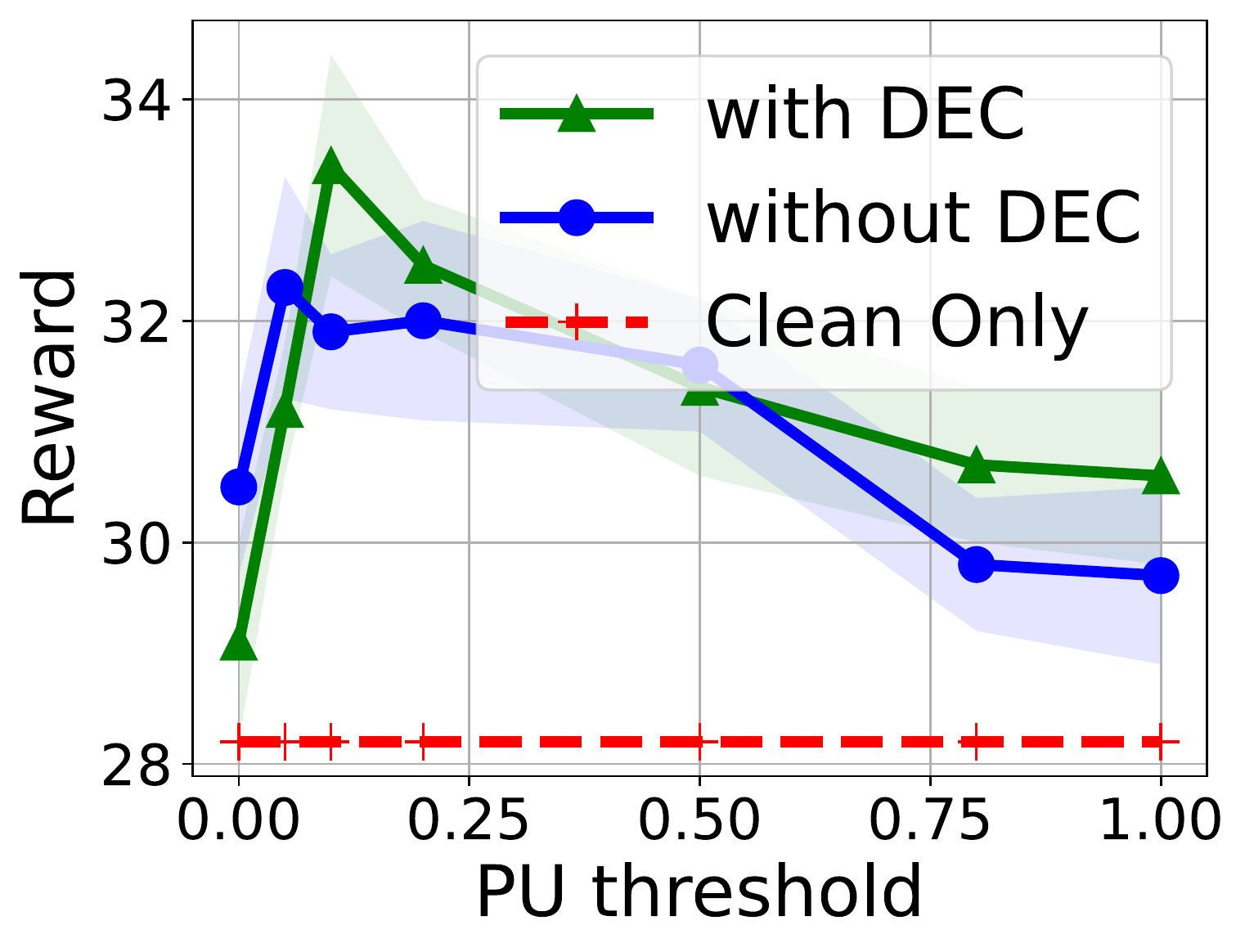}}
    %\subfigure[FourRoom, macroF]{\includegraphics[width=0.32\linewidth]{figs/PU thresholdfourroom_setting0.pdf}}
    \subfigure[Mimic-IV, macroAUC]{\includegraphics[width=0.48\linewidth]{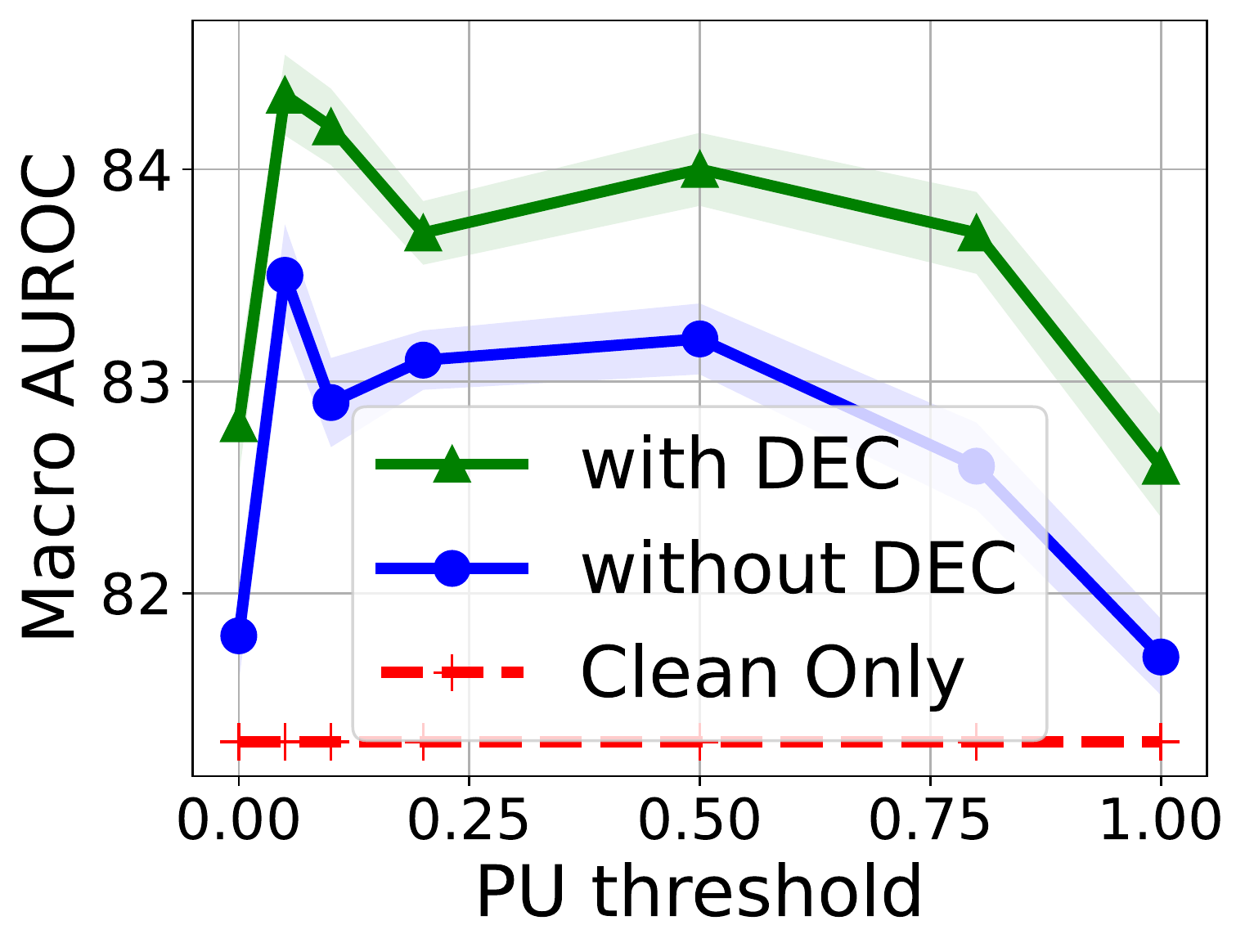}}
    \vskip -1.5em
    \caption{Analysis on PU learning by varying the threshold $\epsilon$ for pseudo optimality estimation. }
    \vskip -1.5em
    \label{fig:PU_sensitivity}
\end{figure}
\begin{figure}
    \centering
    \subfigure[FourRoom, Reward]{\includegraphics[width=0.48\linewidth]{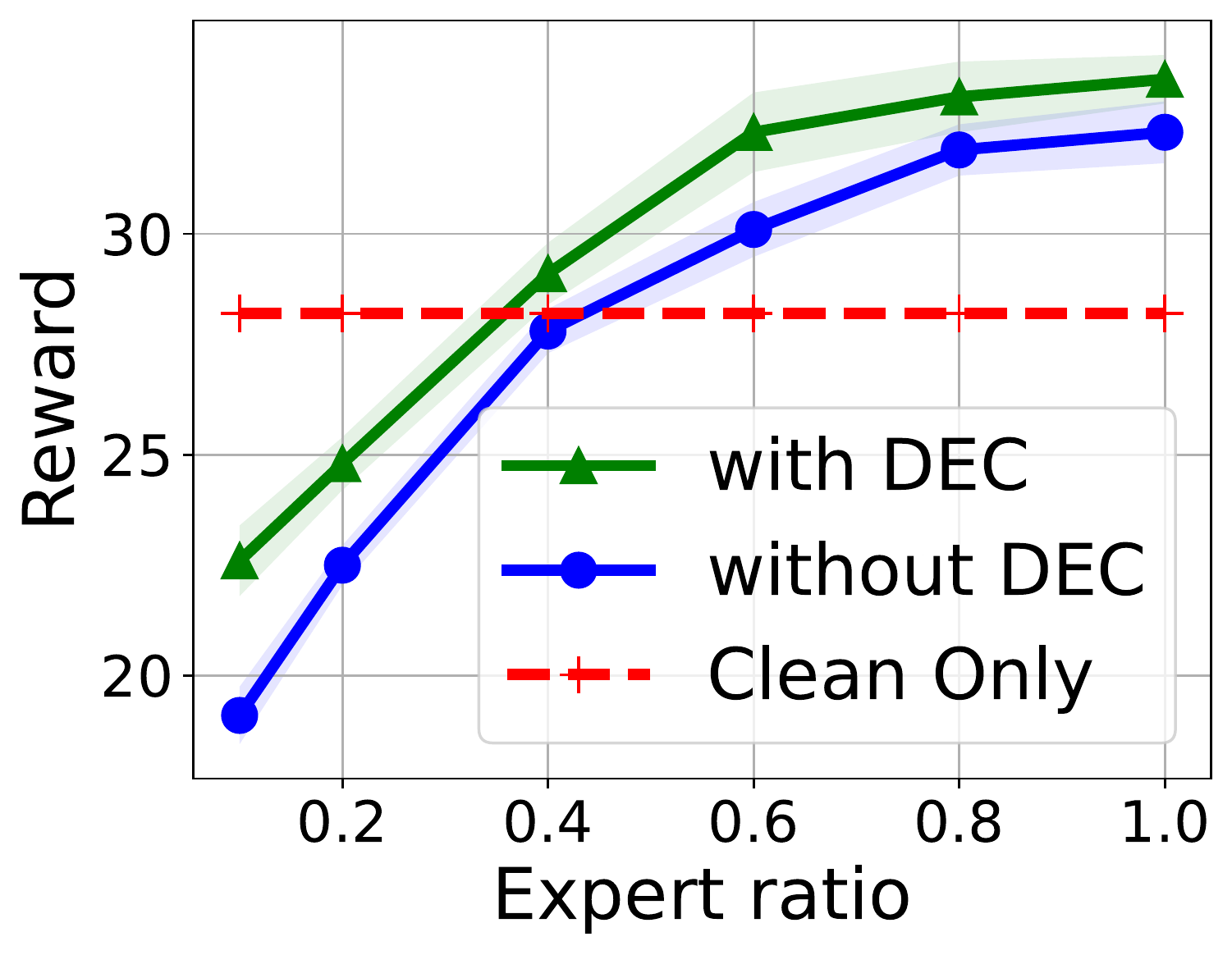}}
    %\subfigure[FourRoom, macroF]{\includegraphics[width=0.32\linewidth]{figs/Expert ratiofourroom_setting0.pdf}}
    \subfigure[Mimic-IV, macroAUC]{\includegraphics[width=0.48\linewidth]{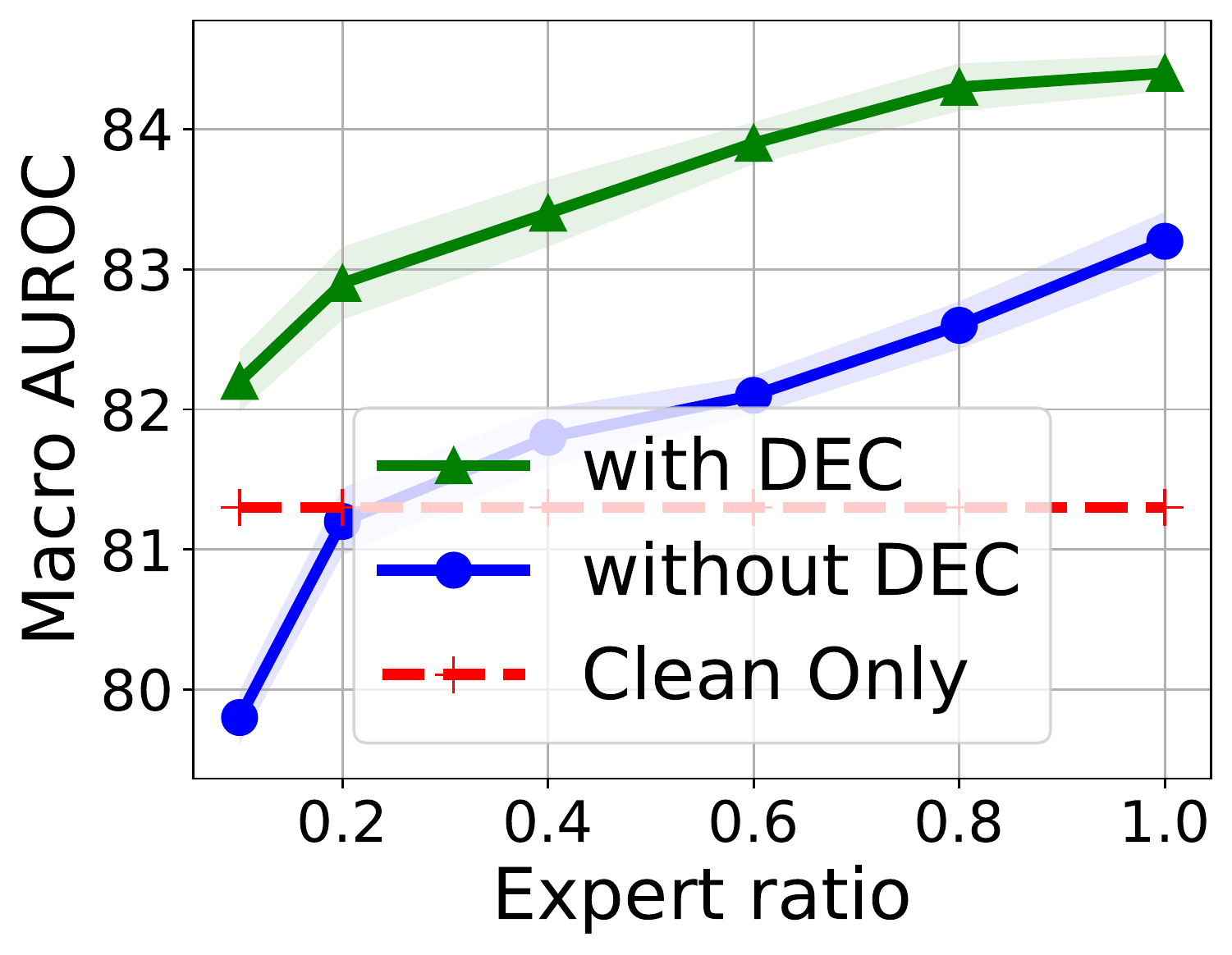}}
    \vskip -1.5em
    \caption{Analysis on sizes of $\mathcal{D}^{clean}$ by taking different ratios of training examples.  }
    \vskip -1.5em
    \label{fig:clean_sensitivity}
\end{figure}

From Fig.~\ref{fig:mi_sensitivity}, it can be observed that when $\lambda$ is $0$, in which case mutual information-based regularization is not applied in the skill discovery phase, the framework is ineffective to extract useful knowledge from noisy demonstrations. Generally, increasing $\lambda$ would improve model performance within the range $[0,1]$. When $\lambda$ becomes larger than $2$, a negative correlation with the performance can be observed. This phenomenon may arise that mutual information loss $\mathcal{L}_{mi}$ dominates the skill discovery process, resulting in learned skills performing poorly w.r.t imitation learning loss $\mathcal{L}_{imi}$.  Generally, setting $\lambda$ within $[0.8, 1.5]$ shows strong performance for both datasets and outperforms the baseline with only clean demonstrations with a clear margin, and applying DEC can further achieve constant improvements. 

\textbf{Utilizing $\mathcal{D}^{noisy}$ With PU Learning.} To utilize both $\mathcal{D}^{clean}$ and $\mathcal{D}^{noisy}$ by discovering skills of varying qualities, we incorporate PU learning to augment the skill discovery together with the DEC mechanism. In particular, a hyper-parameter $\epsilon$ is introduced as the threshold of estimated optimality scores, as shown in Sec.~\ref{sec:DEC&PU}.  In this part to answer RQ2, we evaluate model performance with threshold $\epsilon$ varying in $\{0, 0.1, 0.2, 0.4, 0.5,\}$. A larger $\epsilon$ indicates fewer transitions would be identified as trustworthy to be positive/negative, while a smaller $\epsilon$ would result in utilizing more of obtained optimality scores from PU learning but may also submit to more noises meanwhile.

Experiments are conducted with other configurations unchanged and results are visualized in Fig.~\ref{fig:PU_sensitivity}. Again, we implement two variants w/o the DEC part, and present the base method trained only on $\mathcal{D}^{clean}$ for a comparison. It is shown that for dataset FourRoom, $\epsilon$ is recommended to be set around $0.1$. For Mimic-IV, {\method} performs relatively stable with $\epsilon$ within the range $[0.1, 0.5]$.  Setting the threshold too low would introduce more noise while setting it too high would result in information loss. Furthermore, setting $\epsilon$ too low or too high would also reduce the significance of introducing Deep Embedding Clustering in positive/negative pair selection.

\textbf{Influence of $\mathcal{D}^{clean}$ size}
%The benefits of introducing $\mathcal{D}^{noisy}$ to augment the training data would be largely affected by the number of expert demonstrations available. To analyze its influence, we conduct a sensitivity analysis and show the results in Appendix.~\ref{ap:sensitivity}. It is observed when noisy demonstrations are available, our proposed {\method} can outperform its clean version (trained on full $\mathcal{D}^{clean}$) with less than half of the number of clean demonstrations.
The benefits of introducing $\mathcal{D}^{noisy}$ to augment the training data would be largely affected by the number of expert demonstrations available. To examine the improvement of {\method} in such scenarios, in this part we only take a subset of training demonstrations from $\mathcal{D}^{clean}$ as available. For the training set of expert demonstrations, we further vary its availability ratio in $\{0.1, 0.2, 0.4, 0.6, 0.8, 1\}$. All other hyper-parameters remain unchanged, and we report the results in Fig.~\ref{fig:clean_sensitivity}. It can be observed that the model performance would increase with the expert demonstration ratio, and the DEC part shows a constant improvement. Furthermore, it can be observed that our proposed {\method} can outperform the baseline (using only the clean demonstration set $\mathcal{D}^{clean}$) with half the number of clean demonstrations.

\begin{figure}
    \centering
    \subfigure[FourRoom]{
    \includegraphics[width=0.46\linewidth]{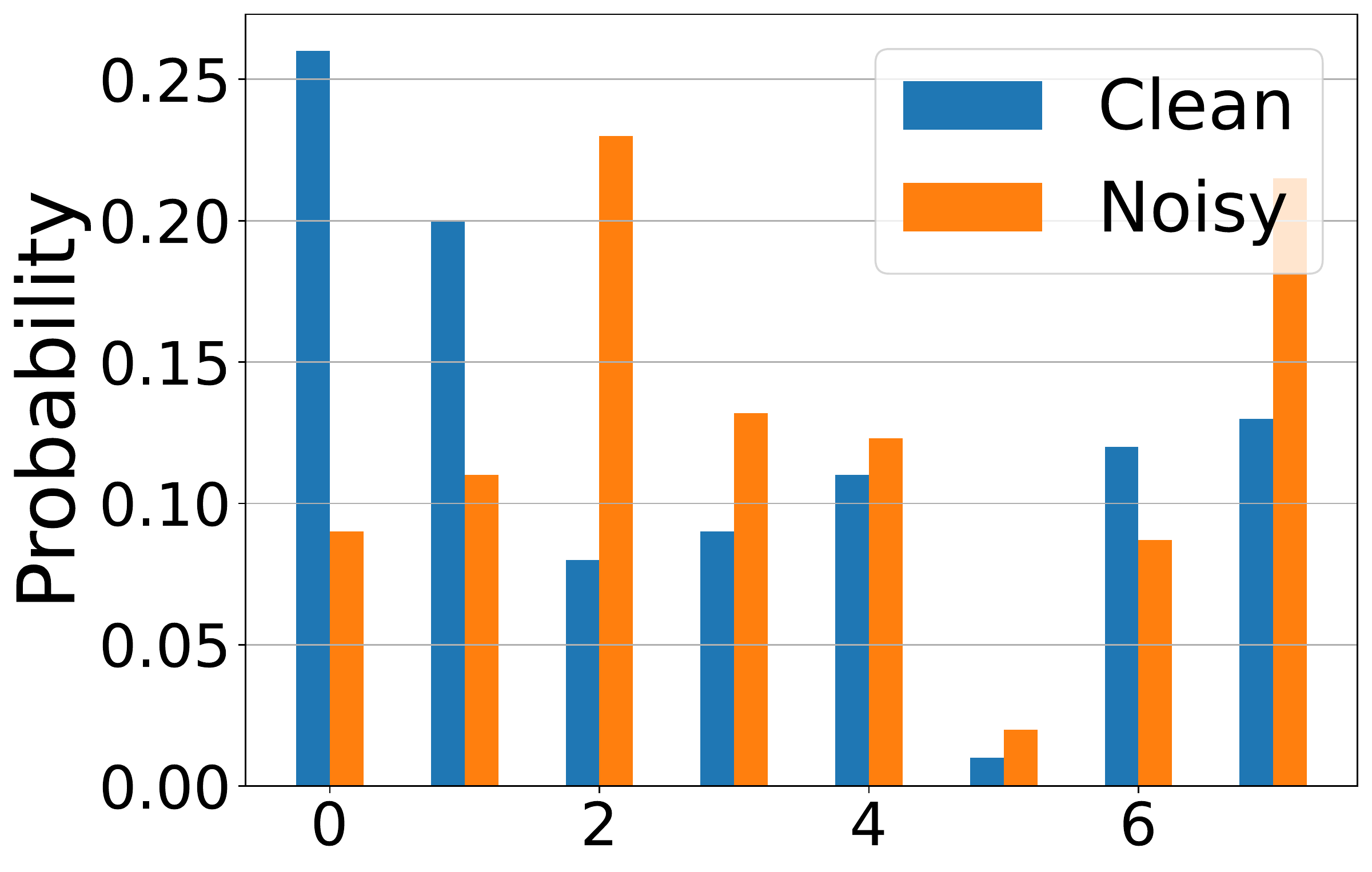}}
    \subfigure[Mimic-IV]{
    \includegraphics[width=0.46\linewidth]{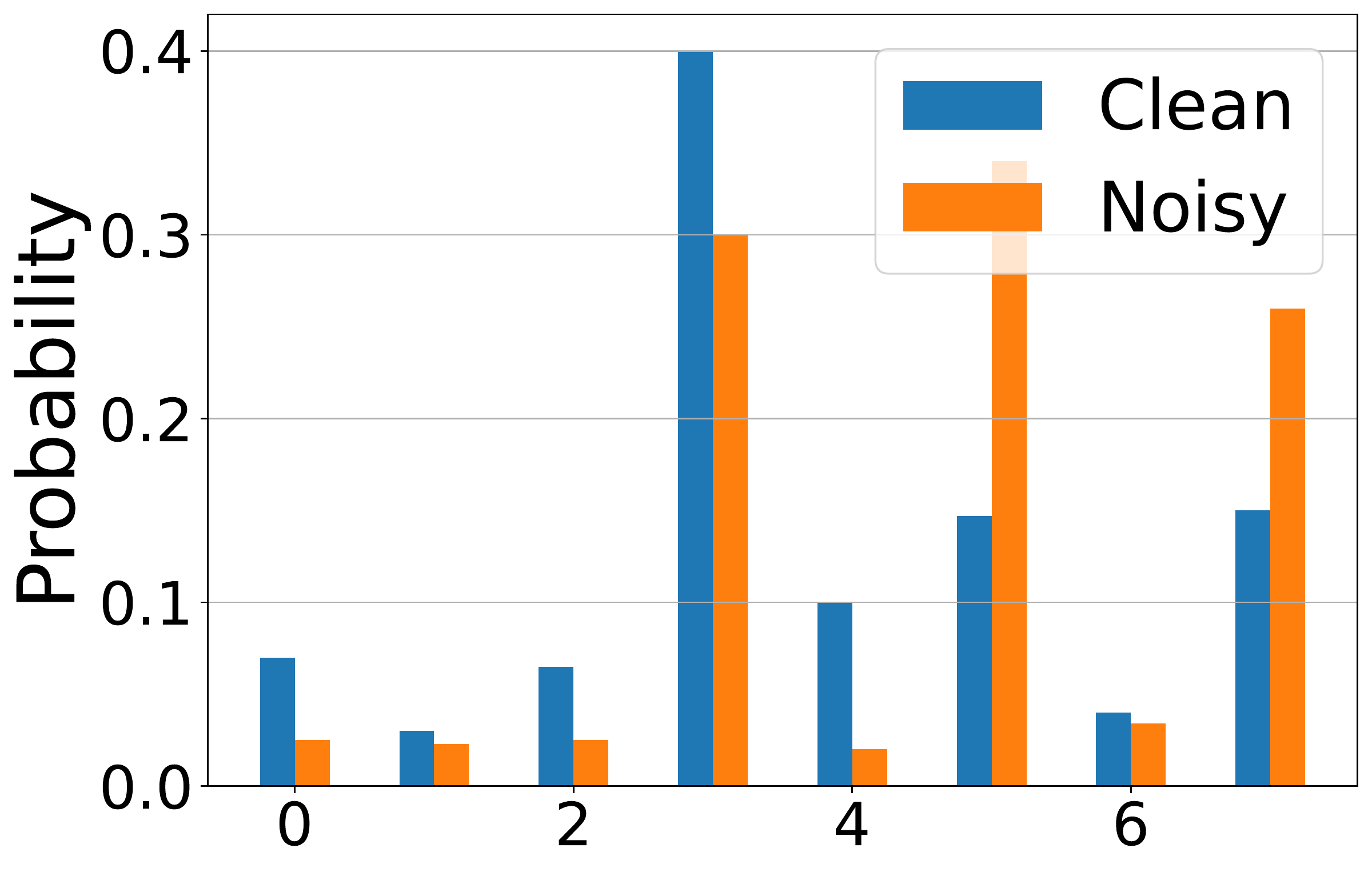}}
    \vskip -1.5em
    \caption{ Comparison of skill selection preference between expert demonstration set and noisy demonstration set on dataset FourRoom. $x$-axis represents the skill variables and $y$-axis is the selection probability.}
    \label{fig:skill_sel}
    \vskip -1.5em
\end{figure}

\textbf{Tendency of Skill Selection Behaviors.} In this part, we want to evaluate the ability of {\method} to identify skills of varying quality by comparing the skill selection behaviors of demonstrations in $\mathcal{D}^{clean}$ and $\mathcal{D}^{noisy}$ statistically. Concretely, we test the trained model on each demonstration and compute the probability of selecting each skill per time step. Skill selection distributions are averaged per demonstration set at each time step in an unweighted manner, and a comparison is made for both the FourRoom dataset (Fig.~\ref{fig:skill_sel}(a)) and the Mimic-IV dataset (Fig.~\ref{fig:skill_sel}(b)).  

In Fig.~\ref{fig:skill_sel}(a) and Fig.~\ref{fig:skill_sel}(b), a clear distribution gap can be observed in the skill selection behavior during imitating demonstrations in $\mathcal{D}^{clean}$ and $\mathcal{D}^{noisy}$. This result verifies the ability of {\method} to extract a skill set of varying optimality, that some are preferred in imitating expert demonstrations while some are preferred by noisy demonstrations. This design also enables effective knowledge sharing across demonstration sets, like the skill $3$ in Fig.~\ref{fig:skill_sel}(b) which are selected frequently by both  $\mathcal{D}^{clean}$ and $\mathcal{D}^{noisy}$. 

\subsection{Case Study}
To provide deeper analysis of the proposed skill discovery process and answer RQ3, we provide a set of case studies on FourRoom and Mimic-IV (Appendix.~\ref{ap:case}) to interpret skills learned by {\method}. 
 Due to the limitation of space, we put results in Appendix.~\ref{ap:case}.

%The analysis of skills is also conducted on the dataset Mimic-IV. 

%% file: secs/conclusion.tex
\section{Conclusion and Future Work}
In this paper, we propose an effective framework {\method} for imitation learning from sub-optimal demonstrations, which is able to utilize both clean and noisy demonstrations simultaneously. Taking a hierarchical architecture, {\method} manages to learn a disentangled skill set of varying optimality and compose an expert neural agent with them. {\method} also offers improved interpretability by analyzing learned skills. Breaking demonstrations down, {\method} can extract knowledge and utilize noisy demonstrations at the segment level, instead of assuming all time steps of the same demonstration to be of the same quality. Experimental results validate the effectiveness of {\method}, and comprehensive analysis is provided for ablation study and examination of learned skills. Note that the core of our method lies upon the discovered skill set. 

As part of future work, we plan to explore the transferability of discovered skills, considering the discovery and manipulation of sub-policy skills  with multiple different objectives.

%% file: secs/appendix.tex
%\section{Training Algorithm}\label{ap:alg}

\section{Datasets}\label{ap:datasets}
    \noindent\textbf{FourRoom.} In FourRoom, the agent needs to navigate in a maze composed of four rooms interconnected by 4 gaps in the walls, and the goal is to reach a specific target position~\cite{gym_minigrid}. The expert demonstration set is constructed by collecting successful trajectories from a pre-trained agent, and the noisy demonstration set contains trajectories that fail to reach the target position (arrive at another position). In total, this dataset contains $2,000$ expert demonstrations (average trajectory length: $23.42$) and $2,000$ noisy demonstrations (average trajectory length: $21.08$).
    
    \iffalse
    \begin{figure}
        \centering
    \subfigure[FourRoom]{\includegraphics[width=0.38\linewidth]{figs/FourRoom.png}}
    \subfigure[DoorKey]{\includegraphics[width=0.37\linewidth]{figs/env_doorkey.jpg}}
    \vskip -1em
        \caption{Illustration of the MiniGrid-FourRoom and MiniGrid-DoorKey environments. The agent starts at a random position, and the objective is to reach the green square. }\label{fig:dataset}
    \vskip -1.5em
    \end{figure}
    \fi
    
    \noindent\textbf{DoorKey.} This environment is a two-room maze, in which the agent must first pick up a key and then unlock the door before getting to the target square in the other room. It is a challenging environment due to the sparsity of rewards. Utilizing a pre-trained agent, the clean and noisy sets are constructed as its generated trajectories of different reward ranges. Concretely, $2,000$ trajectories with rewards higher than $92$ are selected as the clean set (average trajectory length: $9.71$), and $2,000$ trajectories with rewards within $[30, 90]$ are used as the noisy set (average trajectory length: $9.88$).
    
    \noindent\textbf{Mimic-IV.} Mimic-IV contains $43,000$ patients in intensive care units during the year $2001$ and $2012$. Following the procedures in~\cite{wang2022hierarchical}, we extract the \textit{Sepsis} patients from it conforming to the Sepsis-3 criteria~\cite{singer2016third}. For each patient, a trajectory of received medication treatments is extracted (a demonstration), with each time step corresponding to four hours that is the median of the prescription frequency in this dataset~\cite{wang2022hierarchical}. Medication action at each time step is represented as a one-hot vector with a length of $25$. The observation at each time step contains relevant physiological features including both static and temporally dynamic ones, along with the historical medication actions at the previous step. This dataset contains $6,630$ trajectories that the patient is fully recovered (average trajectory length: $12.92$, we use them as clean demonstrations) and $3,573$ trajectories that the patient's health condition deteriorates or is deceased (average trajectory length: $13.08$, used as noisy demonstrations). The objective is to train a neural agent to learn an expert prescription policy from these demonstrations.

    \iffalse
    \begin{figure}
    \centering
    \subfigure[DoorKey]{\includegraphics[width=0.365\linewidth]{figs/env_doorkey.jpg}}
    \subfigure[LavaGap]{\includegraphics[width=0.4\linewidth]{figs/env_lava.jpg}}
    \caption{Illustration of the Doorkey and LavaGap environments. }
    \label{fig:env}
    \end{figure}

    \noindent\textbf{LavaGap.} In this environment, the target region (green square) is put at the opposite corner, and there are deadly lava (orange squares) in the room. Touching the lava would result in termination with a zero reward, and the agent must navigate through a narrow gap to safely arrive at the target. 
    \fi

\section{Baselines}\label{ap:baseline}
In this part, we briefly introduce those baselines compared to in this work. First are those representative imitation learning and treatment recommendation methods:
\begin{itemize}[leftmargin=0.1in]
    \item Behavior Cloning (BC)~\cite{pomerleau1998autonomous}: BC bins the demonstrations into transitions and learns an action-taking policy in a supervised manner. 
    %\item SRL-RNN~\cite{wang2018supervised}: It manually designs a sparse reward function and combines supervised learning and reinforcement learning for dynamic treatment recommendations.
    \item DensityIRL~\cite{wang2020imitation}: Density-based reward modeling conducts a non-parametric estimation of marginal as well as joint distribution of states and actions from provided demonstrations, and computes a reward using the log of those probabilities.
    \item MCE IRL~\cite{ziebart2010modeling}: Its full name is Maximum causal entropy inverse reinforcement learning, whose idea is similar to DensityIRL. It designs an entropy-based approach to estimate causally conditioned probabilities for the estimation of reward function.
    \item AIRL~\cite{fu2018learning}: Adversarial inverse reinforcement learning aims to automatically acquiring a scalable reward function from expert demonstrations by adversarial learning.
    %\item D3Q~\cite{raghu2017continuous}: It is similar to the framework of SRL-RNN and trains the policy model with deep Q-learning.
    \item GAIL~\cite{ho2016generative}: GAIL directly trains a discriminator adversarially to provide the reward signal for policy learning.
    %\item Directed Info-Gail (D-GAIL)~\cite{sharma2018directed}: Directed Info-GAIL extends vanilla GAIL by modeling the generation of expert trajectories as of multiple modes in a graphical model and mimics the expert policy by maximizing the directed information flow.
    %\item SHIL~\cite{wang2022hierarchical}: SHIL designs a hierarchical framework with intermediate subgoals to condition the policy, which is expected to capture the switching from the treatment of one symptom to another.
\end{itemize}
The above methods are all proposed assuming demonstrations to be clean and optimal. We further compare {\method} with two recent strategies that are designed to utilize noisy demonstrations explicitly: ACIL~\cite{wang2020adversarial} and DWBC~\cite{xu2022discriminator}. 
\begin{itemize}[leftmargin=0.1in]
    \item ACIL~\cite{wang2020adversarial}: ACIL extends GAIL by incorporating noisy demonstrations as negative examples and learns the policy model to be distinct from them.
    \item DWBC~\cite{xu2022discriminator}:  It trains a discriminator to evaluate the quality of each demonstration trajectory through adversarial training, and conducts a weighted behavior cloning. 
\end{itemize}

\section{Case Study}\label{ap:case}

FourRoom is a gym environment of four rooms in which the agent needs to navigate and reach the target grid. In this environment, behaviors of the agent are easy to interpret which allows us to directly visualize and evaluate learned skills. To interprete knowledge learned by {\method}, we conduct two sets of studies: 1) distribution of skill selections along clean and noisy demonstrations; 2) action-taking strategies encoded by each concrete skill.

\subsection{Skill Selection Behaviors}
\begin{figure}
    \centering
    \subfigure[Skill selection of    $\mathcal{D}^{clean}$]{\includegraphics[width=0.44\linewidth]{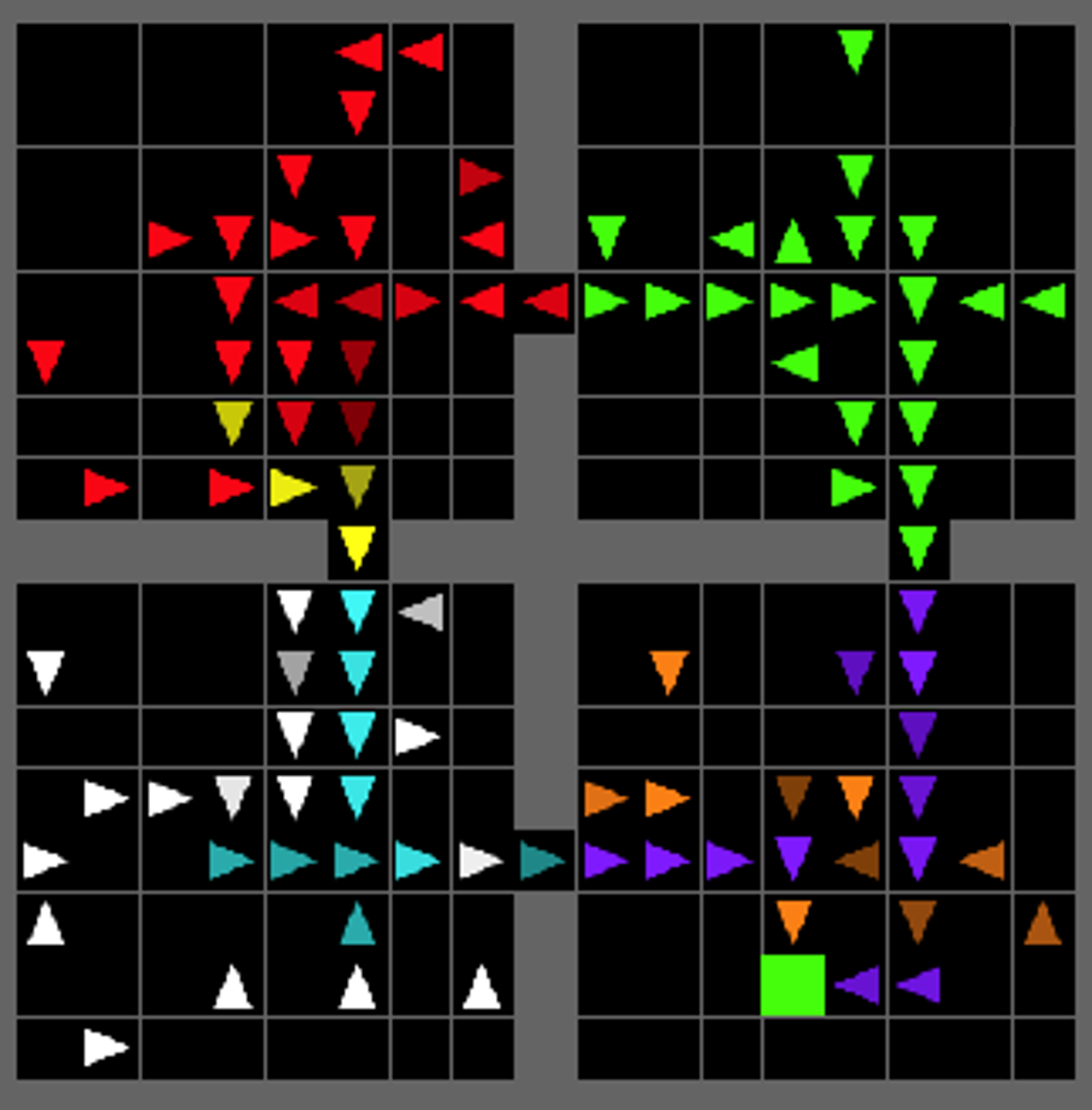}}
    ~~\subfigure[Skill selection of $\mathcal{D}^{noisy}$]{\includegraphics[width=0.44\linewidth]{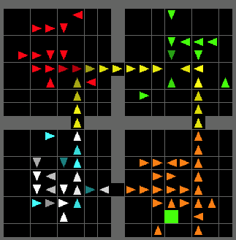}}
    \vskip -1.5em
    \caption{Visualization of skill selection behavior on dataset FourRoom. We test the bi-directional skill selection module on demonstrations sampled from $\mathcal{D}^{clean}$ and $\mathcal{D}^{noisy}$ respectively, and use each color to denote a concrete skill variable. Transparency represents confidence of the selected skill. }
    \label{fig:skill_demon_FourRoom}
    \vskip -1.5em
\end{figure}

First, we analyze skill selection behaviors of the expert demonstration set, and compare them with skill selection behaviors of the noisy demonstration set in Fig.~\ref{fig:skill_demon_FourRoom}. Concretely, we test the skill selection behavior of {\method} on given demonstrations for a case study. $50$ demonstrations are selected from $\mathcal{D}^{clean}$ and $\mathcal{D}^{noisy}$ respectively, and we visualize them in the FourRoom environment as in Fig.~\ref{fig:skill_demon_FourRoom}. Arrow direction represents the majority moving direction of sampled demonstrations at that position, and those that have a high diversity in action selection are discarded. We annotate the selected skills using different colors, and its transparency shows the consistency of skill selection across demonstrations at that position. From the figure, it can be observed that each skill encodes a particular moving strategy inside a region, and some skills are shared by both demonstration sets like the red one. There are also some skills encoding moving strategy specific for one demonstration set, like the purple one in Fig.~\ref{fig:skill_demon_FourRoom}(a) and the yellow one in Fig.~\ref{fig:skill_demon_FourRoom}(b).

\subsection{Behavior Learned by Skills}

\begin{figure*}
    \centering
    \subfigure[Skill with $1st$ preference score by $\mathcal{D}^{clean} $ over $\mathcal{D}^{noisy}$]{\includegraphics[width=0.2\linewidth]{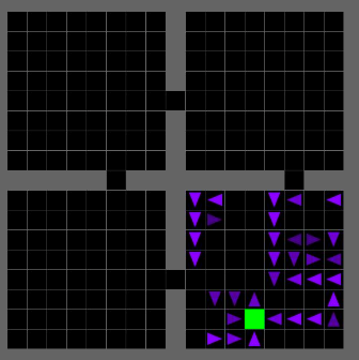}}
    \subfigure[Skill with $2nd$ preference score by $\mathcal{D}^{clean} $ over $\mathcal{D}^{noisy}$]{\includegraphics[width=0.2\linewidth]{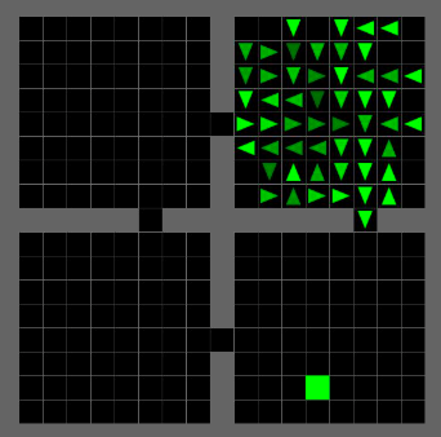}}
    \subfigure[Skill with $3rd$ preference score by $\mathcal{D}^{clean} $ over $\mathcal{D}^{noisy}$]{\includegraphics[width=0.2\linewidth]{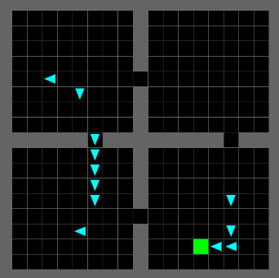}}
    \subfigure[Skill with $4th$ preference score by $\mathcal{D}^{clean} $ over $\mathcal{D}^{noisy}$]{\includegraphics[width=0.2\linewidth]{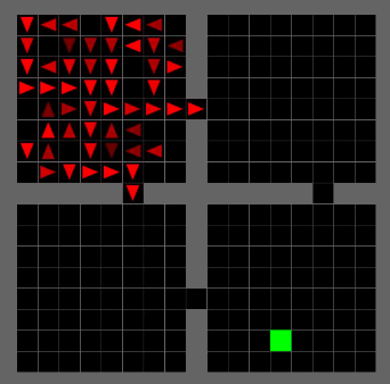}}
    
    \subfigure[Skill with $5th$ preference score by $\mathcal{D}^{clean} $ over $\mathcal{D}^{noisy}$]{\includegraphics[width=0.2\linewidth]{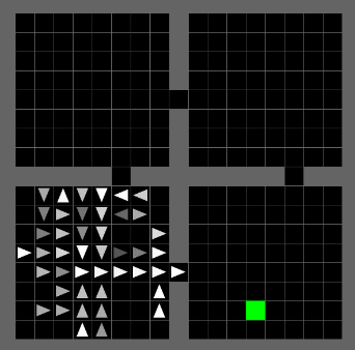}}
    \subfigure[Skill with $6th$ preference score by $\mathcal{D}^{clean} $ over $\mathcal{D}^{noisy}$]{\includegraphics[width=0.2\linewidth]{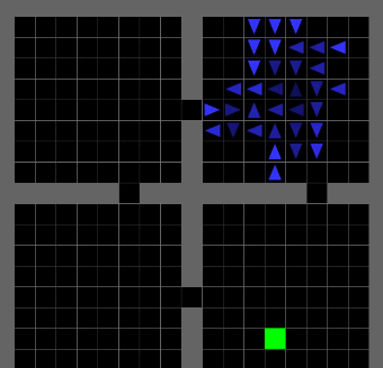}}
    \subfigure[Skill with $7th$ preference score by $\mathcal{D}^{clean} $ over $\mathcal{D}^{noisy}$]{\includegraphics[width=0.2\linewidth]{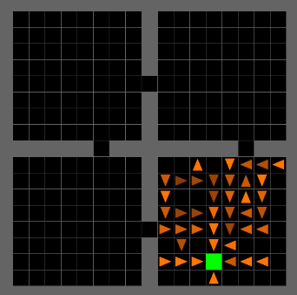}}
    \subfigure[Skill with $8th$ preference score by $\mathcal{D}^{clean} $ over $\mathcal{D}^{noisy}$]{\includegraphics[width=0.2\linewidth]{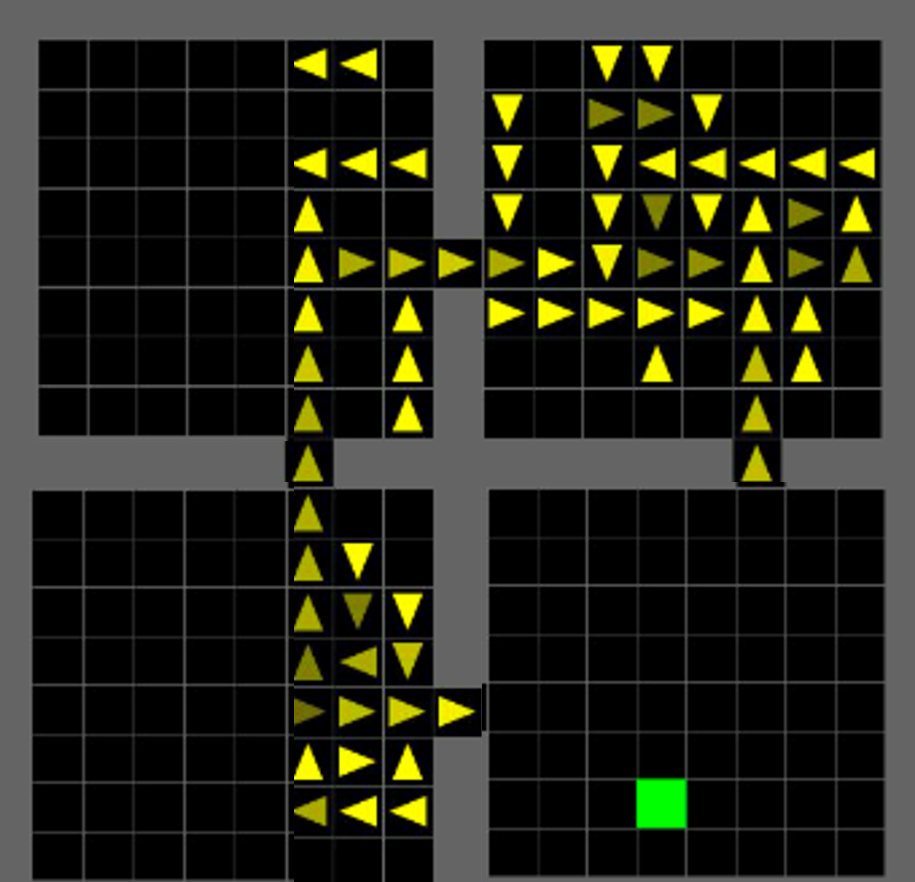}}
    \caption{Visualization of learned skills from dataset FourRoom, after the skill discovery phase. We rank them through comparing their probability of being selected by $\mathcal{D}^{clean}$ over $\mathcal{D}^{noisy}$. For each skill, we randomly run the agent to generate $50$ trajectories with the skill variable fixed. We only visualize the transitions when the pre-fixed skill has the highest probability of being selected to better highlight its encoded behaviors. Arrow at each grid denotes the moving direction, and its transparency represents confidence of the selected action.  }
    \label{fig:skill_vis}
\end{figure*}

\textbf{FourRoom}
Here, we visualize the learned skills in Fig.~\ref{fig:skill_vis}. For each skill, we force {\method} to generate $50$ trajectories with the skill variable fixed. For ease of interpretation, we only visualize the regions in which this skill has a high probability of being selected itself. Arrow direction indicates the majority moving direction, and its transparency shows its consistency. We rank the skills based on their preference by $\mathcal{D}^{clean}$ over $\mathcal{D}^{noisy}$. It is clear to see that: 1) Each skill represents a particular moving strategy within a region. For example, Fig.~\ref{fig:skill_vis}(a) shows the skill of heading to the target region inside the same room, and Fig.~\ref{fig:skill_vis}(c) corresponds to the behavior of going across a pathway in a particular direction. 2) Our {\method} is able to discover separate skills for different strategies of the same region, like comparing Fig.~\ref{fig:skill_vis}(b) to Fig.~\ref{fig:skill_vis}(f). Skill in Fig.~\ref{fig:skill_vis}(b) targets to move to the lower-right room, while that in Fig.~\ref{fig:skill_vis}(f) would stay in this room and head to a particular position.

\begin{figure}[h]
    \centering
    \subfigure[$\lambda=0$]{\includegraphics[width=0.47\linewidth]{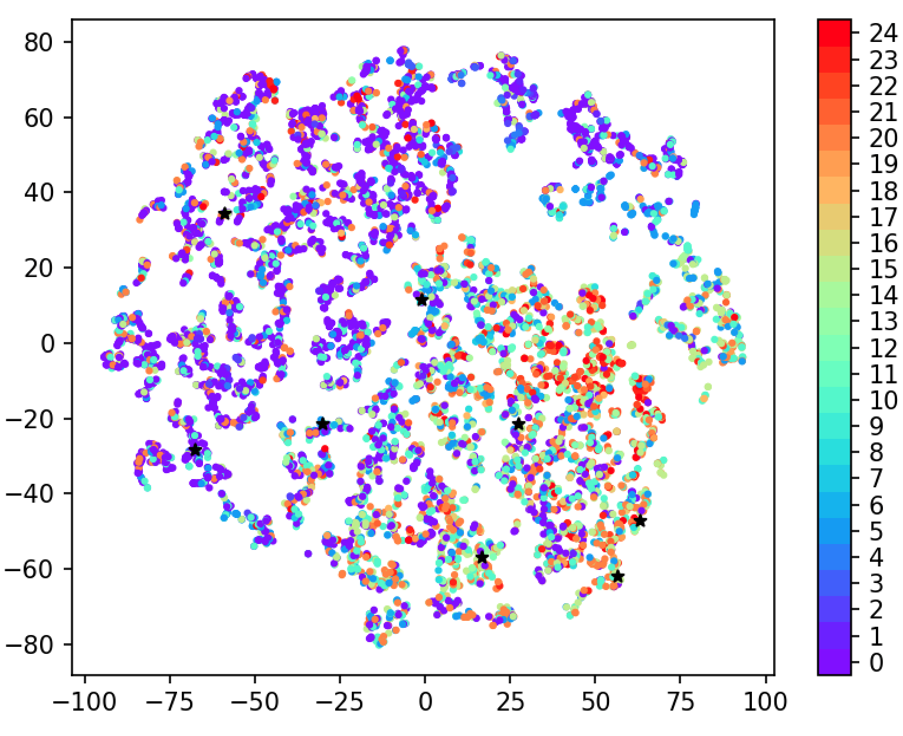}}
    \subfigure[$\lambda=1$]{\includegraphics[width=0.47\linewidth]{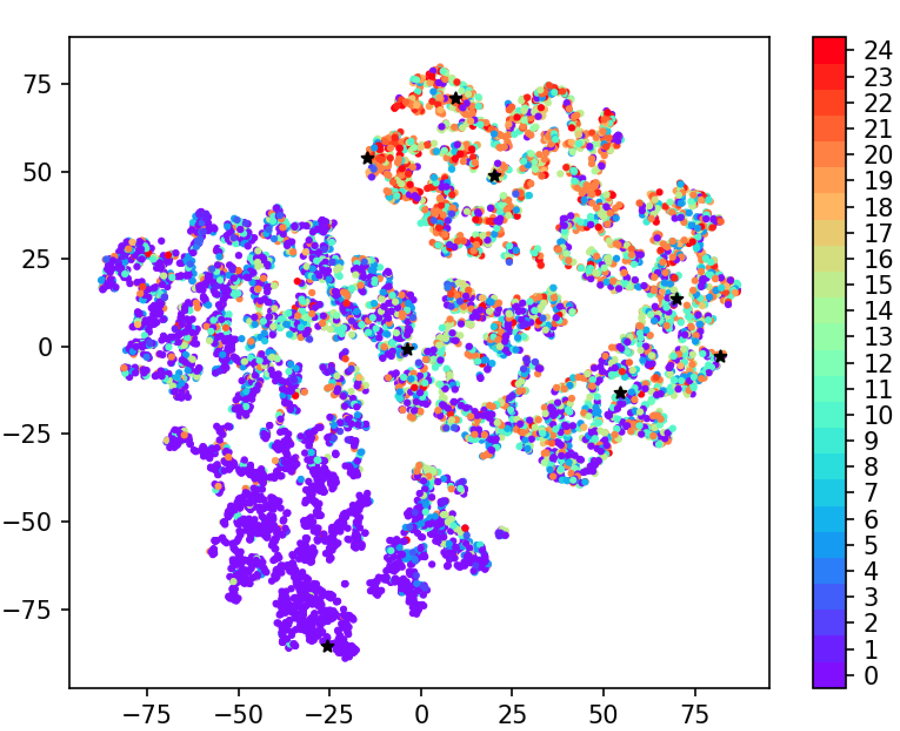}}
    \vskip -1em
    \caption{TSNE Visualization of skill distribution of dataset Mimic-IV. Each color represents a different treatment action, and embedding of skill variables are marked with a black pentagram. In our data split, severity of health conditions would generally increase with a larger treatment class.}
    \label{fig:skill_demon_sepsis}
    \vskip -1em
\end{figure}

\textbf{Mimic-IV}
In the medical dataset Mimic-IV, it is difficult to directly interpret learned skills, as each skill would encode a particular distribution of actions conditioned on observed states. Faced with these difficulties, we conduct the case study from another perspective: by analyzing the distribution of latent embeddings w.r.t skill variables and target actions. Concretely, we draw the TSNE visualization in Fig.~\ref{fig:skill_demon_sepsis}. Each node represents the embedding of a transition extracted by the skill encoding module, black pentagrams denote the embedding of skill variables, and the color represents the selected treatments. Note that in Mimic-IV dataset, a larger treatment label corresponds to a more active medical plan. To further make the comparison, we test two variants of {\method} with the weight of mutual information regularization being set to $1$ and $0$ respectively. Results are shown in Fig.~\ref{fig:skill_demon_sepsis}.
Comparing two figures, it can be observed that skills in Fig.~\ref{fig:skill_demon_sepsis}(b) has a stronger correlation with treatment distributions. Transitions taking similar treatments are clustered better compared to Fig.~\ref{fig:skill_demon_sepsis}(b).